%% file: aaai2026.tex
\documentclass[letterpaper]{article} 
\usepackage[preprint]{aaai2027}  

\usepackage[most]{tcolorbox}
\usepackage{booktabs}
\usepackage{pifont}
\usepackage{amsthm,amsmath,amssymb}
\usepackage{multirow} 
\usepackage{enumitem}
\usepackage{fontawesome}
\usepackage{array}
\usepackage{colortbl} 
\usepackage{xcolor} 
\usepackage{subcaption}

\newcommand{\cmark}{\textcolor{green!70!black}{\ding{51}}}
\newcommand{\xmark}{\textcolor{red!70!black}{\ding{55}}}
\newcolumntype{C}[1]{>{\centering\arraybackslash}p{#1}}
\newcolumntype{L}[1]{>{\raggedright\arraybackslash}p{#1}}

\usepackage{makecell}   
\providecommand{\err}[1]{{\scriptsize\color{gray}$\pm$#1}}
\usepackage{times}  
\usepackage{helvet}  
\usepackage{courier}  
\usepackage[hyphens]{url}  
\usepackage{graphicx} 
\usepackage{natbib}  
\usepackage{caption} 
\usepackage{algorithm}
\usepackage{algorithmic}
\theoremstyle{definition}

\usepackage{newfloat}
\usepackage{listings}
\DeclareCaptionStyle{ruled}{labelfont=normalfont,labelsep=colon,strut=off} 
\floatstyle{ruled}
\newfloat{listing}{tb}{lst}{}
\floatname{listing}{Listing}

\title{EnvCraft: Synthesizing Executable Environments in Agentic RL for Claw-like Agent }

\author{
    Yirong Zeng\textsuperscript{\rm 1,3},
    Shen You\textsuperscript{\rm 1 3},
    Jinhang Feng\textsuperscript{\rm 2 3},
    Yufei Liu\textsuperscript{\rm 2 3},
    Xiao Ding\textsuperscript{\rm 1}\thanks{Corresponding author: xding@ir.hit.edu.cn}, \\
    Yutai Hou\textsuperscript{\rm 3},
    Hao Cong\textsuperscript{\rm 4 3},
    Yuxian Wang\textsuperscript{\rm 3},
    Wu Ning\textsuperscript{\rm 3},
    Wang Xu\textsuperscript{\rm 3},
    Bibo Cai\textsuperscript{\rm 1}
}
\affiliations{
    \textsuperscript{\rm 1}Harbin Institute of Technology, SCIR Lab, 
    \textsuperscript{\rm 2}Peking University, \\
    \textsuperscript{\rm 3}Huawei Technologies Co., Ltd, 
    \textsuperscript{\rm 4}Tsinghua University\\
    $\dagger$ Work done while interning at Huawei.
}

\begin{document}

\maketitle

\begin{abstract}
The paradigm of LLMs has rapidly shifted from passive language interfaces to autonomous Claw-like agents that execute long-horizon tasks across stateful workspaces. 
While Agentic Reinforcement Learning (Agentic RL) provides a promising path to optimize these agents, its scaling is heavily bottlenecked by the severe scarcity of interactive training environments. 
Existing synthetic environments are strictly limited to tool-calling endpoints, rendering them insufficient for accommodating the end-to-end real-world demands of claw-like agents.
To bridge this gap, we introduce EnvCraft, an automated framework for synthesizing executable environments and scalable training data.
Specifically, EnvCraft employs an environment synthesis engine to build sandbox-isolated workspaces, alongside a topology-aware data generation engine to produce coherent task trajectories. 
Overall, we synthesize 139 interactive environments comprising approximately 20K complex tasks for Agentic RL training.
Experiments on Qwen3/3.5 models (8B–32B) show that our method yields gains of up to +11.9\% on Claw-style benchmarks and +8.0\% on general tool-use benchmarks, with concurrent reductions in inference token cost. 
The results confirm that synthesized executable environments provide robust and generalizable learning signals for training.


\end{abstract}

\begin{links}
    \link{Code}{https://github.com/zeng-yirong/EnvCraft}
    \link{Datasets}{https://huggingface.co/EnvCraft}
\end{links}

\section{Introduction}

AI agents have evolved from passive conversational interfaces into autonomous actors operating in real-world systems. 
A notable embodiment of this evolution is the class of claw-like agents, designed for efficient execution within system-level harnesses.
These harnesses, such as OpenClaw~\citep{openclaw}, NanoClaw~\citep{nanoclaw2026}, and Hermes-Agent~\citep{nousresearch2026hermes}, function as persistent digital assistants that manage long-horizon tasks across operating systems, file systems, databases, and terminal shells. 
Unlike traditional conversational bots, claw-like agents possess active execution capabilities that directly mutate the environment, producing real-world consequences.
Recent works explore \textit{Agentic RL}, which enables agents to iteratively refine their policies, recover from execution errors, and generalize to unseen scenarios.

\begin{figure}[t]
    \small
    \centering
    \includegraphics[width=1.0\linewidth]{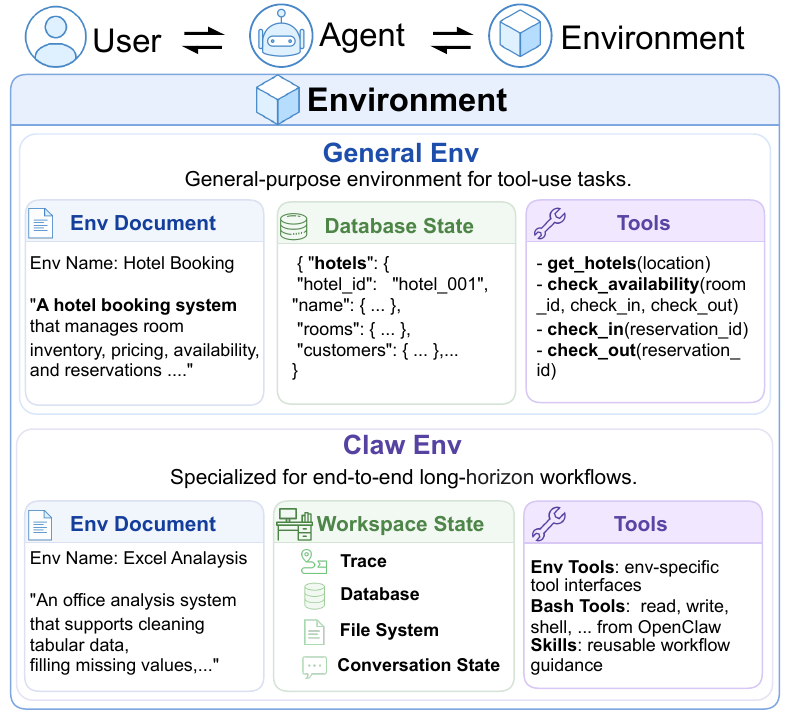} 
    \caption{
      An illustration of the agent interaction loop in RL.
      Environment fundamentally comprises environment documentation, explicit state definitions, and a suite of tool interfaces.
    }
  \label{fig:intro}
\end{figure}

Agentic RL frameworks inherently rely on the interaction among users, agents, and environments, as shown in Figure \ref{fig:intro}.
The environment is an executable sandbox that supports tool execution, state maintenance, and result feedback, 
while an executable environment is foundational to this paradigm, scaling end-to-end Agentic RL training critically requires environments to be low-latency, high-fidelity, and cost-effective.
Currently, environments meeting these rigorous criteria remain scarce, limited primarily to narrow domains such as code interpreters \citep{feng2025retool} and local-retrieval tools \citep{jin2025search}. 
This scarcity becomes particularly acute for Claw-like agents, which demand stateful workspaces and long-horizon task loops, requirements that far exceed the complexity of tool-use endpoints. 
Consequently, the lack of scalable, interactive, and executable environments stands as the key bottleneck for advancing Agentic RL at the system level.

To address this bottleneck, recent studies have explored two main paradigms for generalist environment construction: 
\textit{simulated environments}, which leverage LLMs to mock tool responses \citep{chen2025scaling, li2025simulating, li2026word} but suffer from hallucination and high costs; 
and \textit{synthetic environments}, which auto-generate executable code sandboxes for better scalability \citep{song2026envscaler, xu2026envfactory}.
However, contemporary synthetic methods are strictly limited to application-level, generalist \textit{tool-use agents} by focusing on isolated, tool-use endpoints.
They fail to support system-level \textit{Claw-like agents}, which inherently demand cross-environment execution, stateful workspaces, and asynchronous task loops.
To bridge this critical gap, we present EnvCraft, an automated executable environment synthesis framework tailored specifically for Claw-like agent training while also supporting general-purpose tool-use agents. 
Table \ref{tab:compare} summarizes the comparison results.

\begin{table*}[t]
    \centering
    \begin{tabular}{l |rrcccc}
        \toprule
        \textbf{Environment} & \textbf{\# Envs} & \textbf{\# Tasks} & \textbf{Claw Support} & \textbf{Env Domain} & \textbf{Stateful} & \textbf{Reward} \\ \midrule
        \textbf{AutoForge}\citep{cai2025autoforge} & 10 & 1,078 & \xmark & Consumer Web & DB Schema & LLM Judge \\
        \textbf{EnvScaler}\citep{song2026envscaler} & 191 & 9,000 & \xmark & General & Class Attr. & Rule Check \\
        \textbf{ScaleEnv} \citep{tu2026scaleenv} & 16 & 2,560 & \xmark & General & DB Schema & Rule Check \\
        \textbf{AWM}\citep{wang2026agent} & {1,000} & 10,000 & \xmark & General & DB Schema  & Code \& LLM \\
        \textbf{EnvFactory}\citep{xu2026envfactory} & 85 & 2,575 & \xmark & General & DB Schema & Code Check \\ 
        \midrule
        \rowcolor{purple!10} 
        \textbf{EnvCraft (Ours)} & 139 & \textbf{19,777} & \cmark & {General} & Workspace & Code Check \\
         \bottomrule
    \end{tabular}
    \caption{Comparison of synthetic executable environments for Agentic RL. 
    Unlike existing works limited to generalist tool endpoints, {EnvCraft} provides stateful, workspace-level infrastructure for training \textit{Claw-like agents} with the largest task scale.}
    \label{tab:compare}
\end{table*}

Specifically, \textbf{EnvCraft} operates through two complementary engines: environment synthesis and data generation.
For environment synthesis, we curate 41 general tool-use and 42 Claw-specific scenarios, 
which LLMs expand into detailed environment specifications and tool interface definitions. 
Guided by nine interaction archetypes that prescribe control-flow and state-management patterns, the synthesizer generates fully executable, sandbox-isolated environments. 
For data generation, we introduce a topology-aware sampling strategy to produce high-quality RL tasks comprising user intents, initial states, and verifiers. 
We first build intra- and cross-environment tool dependencies as a bi-level directed graph, from which weighted random walks yield semantically coherent and logically valid tool chains. 
Conditioned on these chains, LLMs reverse-engineer multi-turn user intents that expose information incrementally, while simultaneously generating deterministic verification scripts that check post-execution workspace states to supply reward signals.
Using \textbf{EnvCraft}, we synthesized 139 interactive environments and 19,777 tasks spanning both Claw-specific and general tool-use task.


Extensive experiments on Qwen3-8/32B and Qwen3.5-9B backbones validate EnvCraft's effectiveness via direct RL. 
On Claw-style benchmarks, Claw-specific training yields gains of up to +11.9 points on PinchBench and +11.4 on Claw-Eval, while simultaneously reducing per-task token cost by up to 35\%, indicating more efficient reasoning paths. 
Critically, these performance gains transfer robustly to disjoint general-purpose tool-use benchmarks (BFCL-v3, $\tau^2$-bench), confirming that the learned policies generalize beyond the training distribution rather than merely overfitting to synthesized environments.

\section{Preliminaries}

\subsection{Problem Setup: Agentic Interaction}

\subsubsection{Environment Formulation}
Formally, we conceptualize an executable agentic environment $E$ as a decoupled quadruple that bridges reinforcement learning abstractions with LLM-based interaction:
\begin{equation}
E = \langle \mathcal{D}_{\text{doc}}, \mathcal{I}_{\text{tool}}, \mathcal{S}, \mathcal{T} \rangle
\end{equation}
where each component is structurally specified as follows:
\begin{itemize}
    \item \textbf{Env Documentation} ($\mathcal{D}_{\text{doc}}$):
    A natural-language description encapsulating the global system rules, operational constraints, and safety boundaries of the environment.
    \item \textbf{Tool Interfaces} ($\mathcal{I}_{\text{tool}}$):
    Names, parameters, and descriptions of all tools exposed to the agent, which define the interface for agent–environment interaction.
    \item \textbf{Environment States} ($\mathcal{S}$): The persistent, observable state space, capturing real-world side-effects: $\mathcal{S}_t = \{ s_{\text{data}}, s_{\text{runtime}} \}$, where $s_{\text{data}}$ represents underlying data states (e.g., file metadata or database schemas) and $s_{\text{runtime}}$ represents temporal execution sessions.
    \item \textbf{Sandbox Transition Function} ($\mathcal{T}$): 
    It executes an agent action $a_t \in \mathcal{I}_{\text{tool}}$ within an isolated sandbox, and deterministically yields the next state and observation: $\mathcal{T}: \mathcal{S}_t \times \mathcal{I}_{\text{tool}} \rightarrow \mathcal{S}_{t+1} \times \mathcal{O}_{t+1}$.
\end{itemize}

\subsubsection{Data Point Formulation}
To facilitate optimization via Agentic RL, we formalize each training data point $\mathcal{P}$ generated within the synthetic environment as a structured tuple:
\begin{equation}
\mathcal{P} = \langle \mathcal{Q}, \mathcal{S}_0, \mathcal{V}_{\text{script}} \rangle
\end{equation}
where $\mathcal{Q}$ represents the user query, and $\mathcal{V}_{\text{script}}$ denotes an automated execution verification script,
$\mathcal{S}_0 \in \mathcal{S}$ denotes the concrete environment initial state that instantiates the runtime workspace prior to agent execution,
$\mathcal{V}_{\text{script}}$ programmatically and dynamically audits the environment state transitions $\mathcal{S}_0 \rightarrow \mathcal{S}_{n}$ (e.g., verifying structural file mutations or executing exact metadata checks). 
This design ensures the derivation of deterministic, programmatic reward signals.

\section{Environmental Synthesis Engines}
Operating in an end-to-end fashion, the environment synthesis engine fully automates the transition from natural-language descriptions to executable environments.

\subsection{Env Scenario Collection}
To establish a robust and comprehensive training foundation, we gathered a vast corpus of environment domains spanning both general tool-use tasks and Claw-specific scenarios. 
These raw samples were curated from the OpenClaw community \citep{steinberger2026openclaw, tencent_skillhub, awesome_openclaw} as well as representative prior works \citep{patil2025bfcl, chen2025acebench, song2026envscaler}. 
After rigorous deduplication and semantic filtering, we retained 12 coarse-grained general tool-use domains and 13 Claw-specific categories. 
At a finer granularity, these categories encompass 42 Claw-specific scenarios and 41 general tool-use scenarios, with specifications provided in the Supplementary Material (SM).

Subsequently, we leveraged LLMs to enrich these skeletal domain profiles. This automated expansion systematically transformed brief natural-language descriptions into granular, document-level environment specifications ($\mathcal{E}_{\text{spec}}$) and declarative tool interface specifications ($\mathcal{T}_{\text{spec}}$). 
Together, these dual specifications establish a holistic contract that precisely dictates what the environment must model and what operational interfaces it should expose.

\begin{figure}[t]
  \centering
  \small
  \begin{subfigure}[b]{0.23\textwidth}
    \centering
    \includegraphics[width=\textwidth]{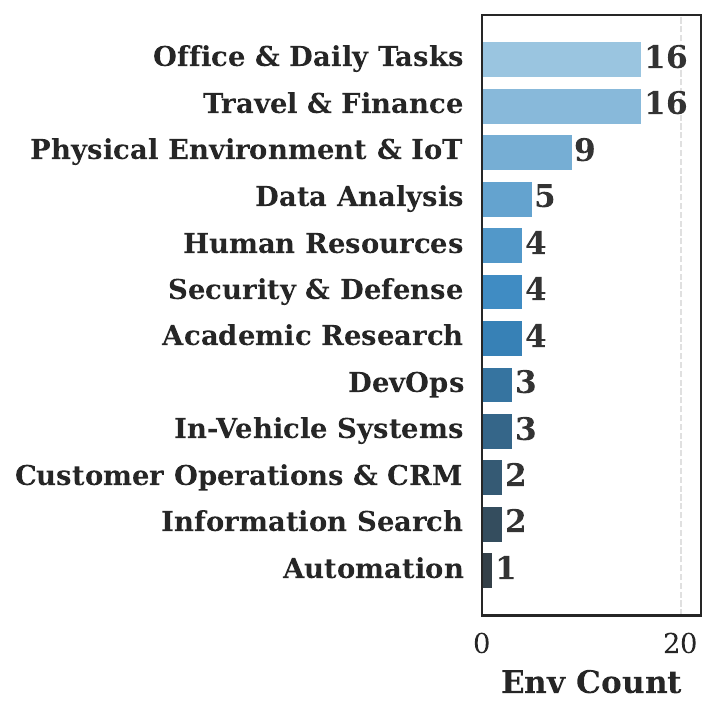}
    \caption{Claw-Specific Scenarios}
  \end{subfigure}
  \begin{subfigure}[b]{0.23\textwidth}
    \centering
    \includegraphics[width=\textwidth]{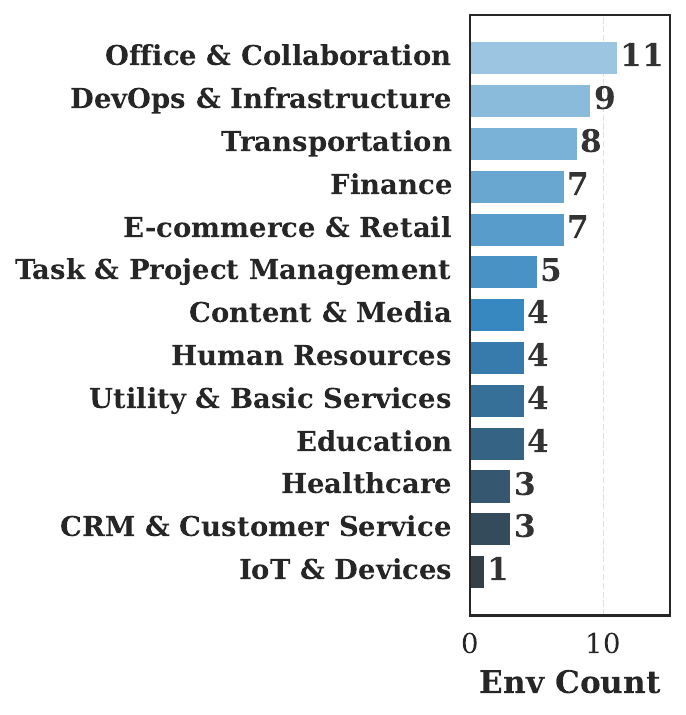}
    \caption{General Tool-Use Scenarios}
  \end{subfigure}
  \caption{Distribution of synthesized environments across coarse-grained scenario categories.}
  \label{fig:22}
\end{figure}

\begin{figure*}[th]
    \centering
    \small
    \includegraphics[width=0.90\textwidth]{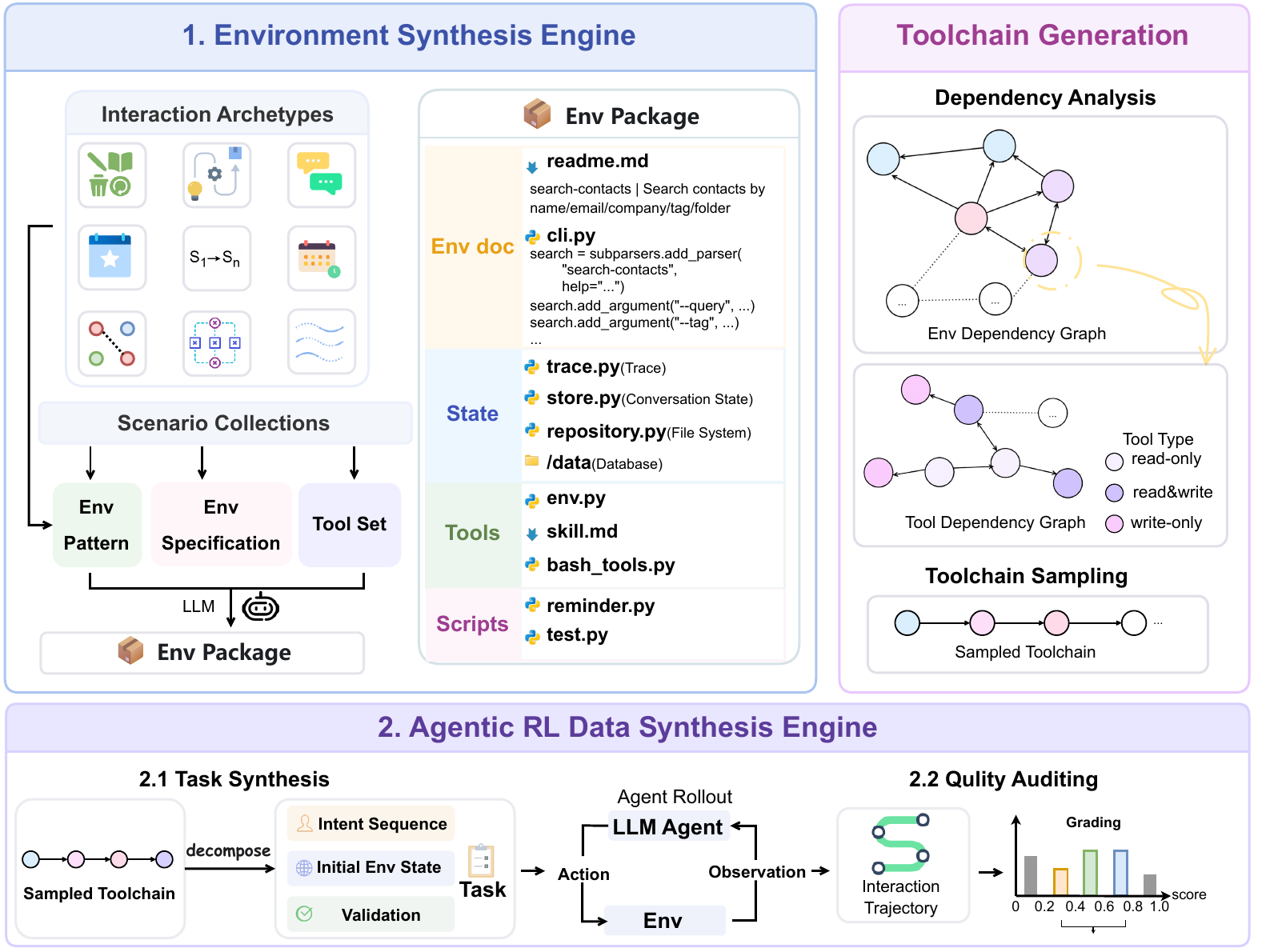} 
    \caption{
    Environment and task synthesis pipeline in EnvCraft. 
    Given natural-language descriptions, the Environment Synthesis Engine generates executable sandboxes (state + tools + docs);
    the Data Synthesis Engine samples tool chains and synthesizes RL training tasks (intent sequence + initial state + verifier).
    }
    \label{fig:overall}
\end{figure*}

\subsection{Environment Sandbox Synthesis}
To generate executable and structurally standard environment sandboxes from high-level natural language descriptions, we propose a prototype-driven synthesis framework. 

\subsubsection{Interaction Archetype Design and Matching}
To accommodate the complex interaction requirements in \textsc{OpenClaw} scenarios, we distill software interaction archetypes into nine foundational \textit{prototypes}. 
Each interaction archetype functions as a structured prompt template, combining abstract design specifications with concrete reference implementations and architectural blueprints.
These nine prototypes cover distinct control-flow and state-management paradigms:
(1) {transactional workspace},
(2) {cross-tool data pipelines},
(3) {dynamic human-agent interaction},
(4) {asynchronous event triggering},
(5) {constraint-guided state machine},
(6) {multi-resource temporal scheduling},
(7) {multi-criteria bilateral matching},
(8) {distributed workflow orchestration}, and
(9) {fault-tolerant stream processing}.
The complete interaction archetype specifications are provided in SM.

Each interaction archetype serves as a design contract, pairing abstract specifications with reference implementations for a target interaction paradigm. 
To ground LLM generation, the framework retrieves relevant prototypes by scoring keyword overlap with the environment specification ($\mathcal{E}_{\text{spec}}$). 
The candidate prototypes are then ranked, and up to the top two are selected as in-context architectural anchors.

\subsubsection{Specification-Guided Environment Generation}
The environment synthesis engine ingests three structured inputs:
\begin{enumerate}[leftmargin=*]
    \item \textbf{Interaction Archetype Specifications ($\mathcal{P}_{\text{proto}}$):} Provide concrete design prototypes, reference implementations.
    \item \textbf{Environment Specification ($\mathcal{E}_{\text{spec}}$):} Defines domain entity types, state attributes, and business rules governing valid state transitions.
    \item \textbf{Tool Specification ($\mathcal{T}_{\text{spec}}$):} Formulates the API interface surface, specifying function schemas, typed parameters, return formats, and side-effect flags (\textit{e.g.}, read-only vs. write-only).
\end{enumerate}
To synthesize executable sandboxes, the LLM processes a unified structured prompt that seamlessly integrates $\mathcal{P}_{\text{proto}}$, $\mathcal{E}_{\text{spec}}$, and $\mathcal{T}_{\text{spec}}$.
Guided by these reference implementations and structured output specifications, the LLM programmatically generates a fully executable environment class. 
A representative synthesized environment instance is illustrated in Figure~\ref{fig:overall}.
The system prompts and execution schemas are detailed in the SM.

\subsubsection{Claw-Specific Extension}
To accommodate the unique execution demands of Claw-specific scenarios, we introduce two domain-specific enhancements:
\begin{itemize}[leftmargin=*]
    \item \textbf{Built-in Tools:} We inject a system-level Bash execution tool derived from the OpenClaw harness into the environment's default toolset, granting agents direct CLI-level interaction capabilities.
    \item \textbf{Parallel Skill Co-Synthesis:} Concurrently with the environment generation, we prompt the synthesizer to produce a set of domain-tailored \textit{Agent Skills} (reusable workflow guidance in markdown).
\end{itemize}

\subsection{Environment Quality Auditing}
To ensure high fidelity, structural robustness, and runtime stability, all synthesized environments must undergo a rigorous dual-stage automated quality control protocol:

\noindent
(1) \textbf{Static code analysis}. We leverage Abstract Syntax Tree (AST) parsing alongside predefined compliance rules to systematically verify the syntactic correctness of the source code, ensuring that all tool schemas and interface definitions strictly adhere to the designated structural specifications. 

\noindent
(2) \textbf{Dynamic runtime analysis}. We programmatically instantiate each environment from its initial state $\mathcal{S}_0$ and execute predefined test cases.
This process tests the operational validity of all exposed tools $\mathcal{I}_{\text{tool}}$, validating state transition logic ($\mathcal{S}_t \rightarrow \mathcal{S}_{t+1}$) and auditing exception-handling behaviors under edge cases. 
Environments that successfully pass both stages are committed to the final library. 
Conversely, failed environments are routed to a self-healing loop, where an LLM is prompted with execution logs to automatically patch the codebase for re-verification. 

Finally, as shown in Table \ref{tab:env_pass_rates}, quality auditing reduced the 324 raw environments to 139 (42.9\% pass rate), comprising 70 general-tool environments and 69 Claw-specific environments.

\section{Data Synthesis Engines}
This section presents our pipeline for synthesizing executable and verifiable tool-use tasks. It includes tool chain sampling, task construction, and quality auditing.


\subsection{Sampling Tool Chain} 

To ground subsequent task synthesis in valid execution logic, we model tool and environment dependencies as a bi-level directed graph structure. 
At the bottom level, we construct an intra-environment \textit{tool dependency graph} for each sandbox. 
By combining static code analysis with LLM reasoning over tool pre- and post-conditions, we map out logical sequences (e.g., data flows) as weighted directed edges. 
At the top level, to support cross-system workflows, we build a cross-environment \textit{env dependency graph}. 
Here, environments act as nodes, and the LLM establishes weighted edges based on semantic domains and cross-environment data flows.

We generate executable tool chains via a hierarchical weighted random walk over this bi-level structure.
For single-environment tasks, the sampler traverses the tool dependency graph; 
for cross-environment tasks, it first samples a sequence of logically connected environments (typically 2--5) from the env-graph, then performs intra-environment walks within each tool-graph to generate and concatenate prefixed sub-chains. 
To ensure verifiability, we partition all tools into \textit{read tools} (non-state-mutating) and \textit{write tools} (state-altering), and enforce that every sampled chain terminates with a mutation tool. 
This guarantees that each synthesized task ultimately modifies the environment state, enabling programmatic answer verification through state-based evaluation.
Ultimately, these sampled tool chains serve as hard execution skeletons, directly driving the downstream intent reverse-engineering and scenario construction for seed task synthesis.

\begin{table}[th]
    \centering
    \small
    \begin{tabular}{l|rrc}
        \toprule
        \textbf{Env/Data Types} & \textbf{Passed} & \textbf{Raw} & \textbf{Pass Rate} \\
        \midrule
        \rowcolor{gray!10} \multicolumn{3}{l}{\textit{Environment}} &  \\ 
        General Tool Env & 70 & 203 & 34.48\% \\
        Claw  Env & 69 & 121 & 57.02\% \\
        \# Total & 139 & 324 & 42.90\% \\
        \midrule
        \rowcolor{gray!10} \multicolumn{3}{l}{\textit{Data Entry}} &  \\ 
        General Tool Data & 14,215 & 21,975 & 64.69\% \\
        Claw-special Data & 5,562  & 13,489  & 41.23\% \\
        \# Total & 19,777 & 35464 & 55.77\% \\
        \bottomrule
    \end{tabular}
    \caption{Statistics of Environment and Data from EnvCraft, along with pass rate in quality auditing.}
    \label{tab:env_pass_rates}
\end{table}

\subsection{Task Synthesis}
While tool chains specify the operational skeleton, they must be transformed into self-contained training samples. We formulate it as a three-stage synthesis process.

\paragraph{Chain-to-Task Reverse Engineering.}
The LLM analyzes data-flow dependencies along the chain and classifies each parameter as either \textit{user-specified} (grounded literals provided by the user, e.g., order IDs or file paths) or \textit{tool-derived} (intermediate results resolved from preceding tool outputs, e.g., query-returned record IDs).
It then reverse-engineers a natural-language user instruction in which user-specified values are stated explicitly while tool-derived values are referenced only abstractly, never revealing their ground-truth values.
This information-hiding mechanism forces the downstream agent to discover intermediate results through actual tool execution, ensuring task non-speculability.

\paragraph{Intent Decomposition}
Since the above intents are expressed as single composite instructions while real users disclose information incrementally, we decompose each task into a sequence.
Given the composite intent and the tool chain's step structure, the LLM generates a user instruction sequence in which the user reveals information or poses sub-tasks step-by-step, and the agent executes corresponding tools sequentially until the overall objective is fulfilled.

To improve the naturalness of the user intents, we incorporate persona attributes into the decomposition process. 
Specifically, we sample persona profiles from PersonaHub~\citep{ge2024scaling} and compute their semantic similarity with the documentation of all environments. 
For each environment, we retain up to the five most relevant personas (500 in total), and randomly select one during intent decomposition to condition the simulated user's behavior.

\paragraph{Intent-to-Environment Construction.}
This step generates the initial environment state and the corresponding answer verification script. 
Specifically, it takes as input the complete environment and tool source code, together with the decomposed user intent sequence. 
We prompt an LLM to instantiate the initial environment state by populating the workspace with both the entities required for the correct execution path and plausible distractor records. 
In parallel, the LLM generates a deterministic verification script that checks whether the resulting workspace state satisfies the user's intent. 
The final output is a self-contained JSON task record containing the task identifier, user task, initial environment state, and executable verification script.

\subsection{Data Quality Auditing}
To ensure the fidelity and trainability of synthesized data, we apply an auditing protocol:
First, we employ an LLM with a predefined rubric covering instruction clarity, tool chain consistency, and evaluation-function correctness; samples scoring below the acceptance threshold are discarded.
Second, we perform difficulty calibration by running Qwen3.5-27B on each task for eight independent trials and filtering out samples with a 0\% or 100\% pass rate.
Samples whose pass rates fall within an intermediate range, constitute the final training set with a balanced difficulty distribution~\citep{he2025skywork}.

After auditing, we retain 19,777 high-quality samples from 35k raw candidates (55\% pass rate), comprising 14,215 general tool-use (\textbf{EnvCraft-Tool}) and 5,562 Claw-specific instances (\textbf{EnvCraft-Claw}), as shown in Table~\ref{tab:env_pass_rates}.


\section{Experiments}
\label{sec:experiments}
\subsection{Setup}
\paragraph{Implementation.}
We fine-tune Qwen3 and Qwen3.5 series models (ranging from 8B to 32B parameters) using GRPO within the VERL framework~\citep{sheng2024verl}. 
Our primary training strategy focuses on claw-specialized data; we additionally explore a two-stage curriculum that first trains on general tool-use data before fine-tuning on Claw-specific trajectories. 
Training is conducted on 64 GPUs with a rollout batch size of 64, a maximum generation length of 32k tokens, and up to 64 action turns per trajectory. 
We enable the default thinking mode throughout training and inference, and defer all remaining hyperparameter configurations to the Supplementary Material (SM).

\paragraph{Baselines.}
We evaluate EnvCraft on the open-source Qwen3 and Qwen3.5 model series and compare them against representative frontier models, including GPT-5.4~\citep{openai2026gpt54}, Claude Opus 4.6~\citep{anthropic2026claudeopus46}, Kimi K2.5~\citep{moonshot2026kimik25}, and MiniMax M2.7~\citep{minimax2026m27}.
Additionally, we include recent RL data for general tool-use environment synthesis (see Table \ref{tab:compare}): {EnvScaler}~\citep{song2026envscaler}, {ScaleEnv}~\citep{tu2026scaleenv}, {AWM}~\citep{wang2026agent}, and {EnvFactory}~\citep{xu2026envfactory}. 
To ensure a fair comparison, we evaluated models trained using these data in consistent experimental settings.

\begin{table}[t]
  \centering
  \small
  \setlength{\tabcolsep}{3pt}
  \begin{tabular}{@{}l cc cc@{}}
    \toprule
    \multirow{2}{*}{\textbf{Model}}
      & \multicolumn{2}{c}{\textbf{PinchBench}} & \multicolumn{2}{c}{\textbf{ClawEval}} \\
    \cmidrule(lr){2-3} \cmidrule(lr){4-5}
      & Score & Token & Score & Token \\
    \midrule
    \multicolumn{5}{@{}l}{\textit{\textbf{Advanced Models}}} \\
    Claude Opus 4.6 & 69.90 & 12.30 & 80.60 & 20.40 \\
    Kimi-2.5 & 54.60 & 14.00 & 66.60 & 17.80 \\
    GPT 5.4 & 75.70 & 11.50 & 78.30 & 15.30 \\
    
    MiniMax-m2.7 & 65.40 & 11.60 & 71.80 & 17.80 \\
    \bottomrule
    \multicolumn{5}{@{}l}{\textit{\textbf{Main Results}}} \\
    \rowcolor{gray!10} Qwen3-8B & 12.94\err{0.3} & 13.69 & 44.06\err{0.2} & 15.49 \\
    \hspace{2pt} + EnvCraft-Tool & 14.51\err{0.1} & 17.34 & 45.80\err{0.1} & 17.59 \\
    \hspace{2pt} + EnvCraft-Tool-Claw & 24.19\err{0.2} & 11.26 & \textbf{55.63}\err{0.3} & 13.25 \\
    \hspace{2pt} + EnvCraft-Claw & \textbf{24.85}\err{0.1} & 8.84 & 55.42\err{0.2} & 12.44 \\
    \midrule
    \rowcolor{gray!10} 
     Qwen3-32B & 22.92\err{0.4} & 17.34 & 60.35\err{0.3} & 23.16 \\
    \hspace{2pt} + EnvCraft-Claw & \textbf{29.51}\err{0.3} & 14.60 & \textbf{62.78}\err{0.2} & 20.33 \\
    \midrule
    \rowcolor{gray!10} Qwen3.5-9B & 47.34\err{0.1} & 11.50 & 72.22\err{0.1} & 19.49 \\
    \hspace{2pt} + EnvCraft-Claw & \textbf{52.19}\err{0.3} & 10.58 & \textbf{76.30}\err{0.1} & 14.32 \\
    \midrule
  \end{tabular}
    \caption{Main results. Score denotes the average task score (\%), and Token denotes the average token usage per trajectory (K). \err{$\cdot$} denotes the standard deviation over three runs. Bold values indicate the best score within each backbone group.}
  \label{main_results}
\end{table}

\paragraph{Evaluation.}
We evaluate on two categories of benchmarks. 
(1) two Claw-style agent benchmarks: PinchBench~\citep{kiloaiteam2026pinchbench} and ClawEval~\citep{ye2026claw}, which assess end-to-end capabilities in solving real-world, long-horizon tasks. 
We employ the standard OpenClaw harness for these evaluations. 
(2) two widely used multi-turn tool-use benchmarks, BFCL-v3~\citep{patil2025bfcl} and $\tau^2$-bench~\citep{yao2024taubench}, which evaluate core tool-use abilities essential for Claw Agents. 
All benchmarks feature domain-specific environments with tools, requiring the LLM to interact with users and invoke tools via its native function-calling interface to ensure consistency. 
We report the average score over three independent runs for each benchmark.

\begin{table*}[t]
  \centering
  \small
  \setlength{\tabcolsep}{4pt}
  \begin{tabular}{@{}l lllll llll@{}}
    \toprule
    \multirow{2}{*}{\textbf{Model}}
      & \multicolumn{5}{c}{\textbf{BFCL-v3 Multi-Turn}}
      & \multicolumn{4}{c}{\textbf{$\boldsymbol{\tau}^2$-bench}} \\ 
    \cmidrule(lr){2-6} \cmidrule(lr){7-10} 
      & \textbf{Base}
      & \makecell{\textbf{Miss-}\\\textbf{Func}}
      & \makecell{\textbf{Miss-}\\\textbf{Param}}
      & \makecell{\textbf{Long-}\\\textbf{Context}}
      & \textbf{Overall}
      & \textbf{Retail}
      & \textbf{Airline}
      & \textbf{Telecom} 
      & \textbf{Overall} \\ 
    \midrule
    \multicolumn{10}{@{}l}{\textit{\textbf{Advanced Models}}} \\ 
    Gemini-3-Pro-Pre.
      & 64.50 & 60.00 & 54.50 & 64.00 & 60.75
      & 75.90 & 80.50  & 91.00 & 82.47\\ 
    Claude Sonnet 4.5
      & 69.00 & 65.00 & 52.50 & 59.00 & 61.37
      & 72.40  & 72.00 & 84.90 & 76.43\\
    Qwen3-235B
      & 60.00 & 35.00 & 34.00 & 54.00 & 45.75
      & 71.90 & 58.60 & 47.30 & 59.27\\
    Kimi-K2-Instruct
      & 57.50 & 35.00 & 42.00 & 49.00 & 45.88
      & 70.60 & 56.50 & 65.80 & 64.30\\
    \midrule 
    \multicolumn{10}{@{}l}{\textit{\textbf{Main Results}}} \\
    \rowcolor{gray!10}
    Qwen3-8B
      & 50.00\err{0.5} & 44.50\err{0.0} & 34.00\err{0.5} & 28.50\err{0.0} & 39.25\err{0.5}
      & 33.33\err{0.1} & 26.00\err{0.0}  & 21.93\err{0.1} & 27.09\err{0.1}\\ 
    \hspace{4pt}ScaleEnv-8B
      & - & - & - & - & -
      & 50.90 & 37.50  & 27.20 & 38.53\\ 
    \hspace{4pt}EnvScaler
      & 55.50 & 36.00 & 35.00 & 41.00& 41.88
      & 53.62 & 36.00  & - & - \\ 
    \hspace{4pt}EnvFactory
      & - & - & - & -& 49.00
      & 43.86 & 44.00  & 13.16 & 33.67 \\ 
    \hspace{4pt}AWM
      & - & - & - & -& -
      & 41.23 & 38.50  & 23.47 & 34.40 \\ 
    \hspace{4pt}EnvCraft-Tool
      & 58.50\err{1.0} & 50.50\err{0.5} & 39.50\err{0.0} & 36.50\err{0.0} & 46.25\err{1.0}
      & 36.84\err{0.2} & 28.00\err{0.0}  & 24.56\err{0.1} & 29.80\err{0.2}\\ 
    \hspace{4pt}EnvCraft-Claw
      & 55.00\err{0.5} & 52.00\err{0.5} & 40.50\err{0.0} & 36.00\err{0.0} & 45.88\err{0.5}
      & 35.52\err{0.1} & 26.50\err{0.5}  & 23.68\err{0.1} & 28.57\err{0.5}\\ 
    \midrule
    \rowcolor{gray!10}
    Qwen3-32B
      & 57.00\err{0.5} & 50.00\err{0.0} & 38.00\err{1.0} & 37.00\err{0.0} & 45.50\err{1.0}
      & 56.14\err{0.2} & 48.00\err{0.0}  & 26.32\err{0.2} & 43.49\err{0.2}\\ 
    \hspace{4pt}ScaleEnv-32B
      & - & - & - & - & -
      & 63.60 & 48.00  & 30.90 & 47.50\\ 
    \hspace{4pt}EnvCraft-Tool
      & 67.00\err{0.0} & 53.00\err{0.5} & 41.50\err{0.5} & 42.50\err{0.0} & 51.00\err{0.5}
      & 56.63\err{0.2} & 52.00\err{0.5}  & 28.07\err{0.2} & 45.57\err{0.5}\\ 
    \midrule
    \rowcolor{gray!10}
    Qwen3.5-9B
      & 52.00\err{1.0} & 53.50\err{0.0} & 39.50\err{0.0} & 34.00\err{0.5} & 44.75\err{1.0}
      & 35.14\err{0.2} & 32.00\err{0.0}  & 15.79\err{0.2} & 27.64\err{0.2}\\ 
    \hspace{4pt}EnvCraft-Tool
      & 65.50\err{0.0} & 59.00\err{0.5} & 45.00\err{0.0} & 41.50\err{0.5} & 52.75\err{0.5}
      & 40.35\err{0.2} & 42.00\err{0.0}  & 26.50\err{0.3} & 36.28\err{0.3}\\ 
    \bottomrule 
  \end{tabular}
  \caption{Performance comparison of advanced models on multi-turn tool-use benchmarks (BFCL-v3 Multi-Turn and ${\tau}^2$-bench).
  All scores are reported as percentages. 
  ``-'' indicates unreported results in the original paper. 
  }
  \label{general_results}
\end{table*}

\subsection{Main Results}
\paragraph{Main Results on Claw-style Benchmarks.}
Table~\ref{main_results} reports performance on PinchBench and Claw-Eval, together with the average token cost per task.
Training on EnvCraft-generated data consistently improves Qwen models across all scales and families.
Concretely, for Qwen3-8B, Claw-specific training boosts PinchBench from 12.94\% to 24.85\% (+11.91\% ) and Claw-Eval from 44.06\% to 55.42\% (+11.36\%). 
Qwen3-32B achieves gains of 6.59 points on PinchBench and 2.43 points on Claw-Eval. 
Qwen3.5-9B shows consistent gains of 4.85 points on PinchBench and 4.08 points on Claw-Eval. 
These improvements across three backbone configurations, ranging from 8B to 32B parameters, 
demonstrate that EnvCraft's benefits generalize beyond a specific scale or architecture, underscoring the value of synthesized environments as a robust training resource.

Comparing training data compositions, general tool-use data yields only modest gains (+1.57\% on PinchBench, +1.74\% on Claw-Eval), whereas Claw-specific trajectories contribute the majority of improvements (+11.91\% and +11.36\%). 
Sequential Tool-to-Claw training achieves comparable results (+11.25\% and +11.57\%), indicating that domain-aligned executable trajectories are more critical than simply accumulating heterogeneous data. 

\begin{figure}[t]
    \centering
    \begin{subfigure}[b]{0.46\textwidth}
        \centering
        \includegraphics[width=\linewidth]{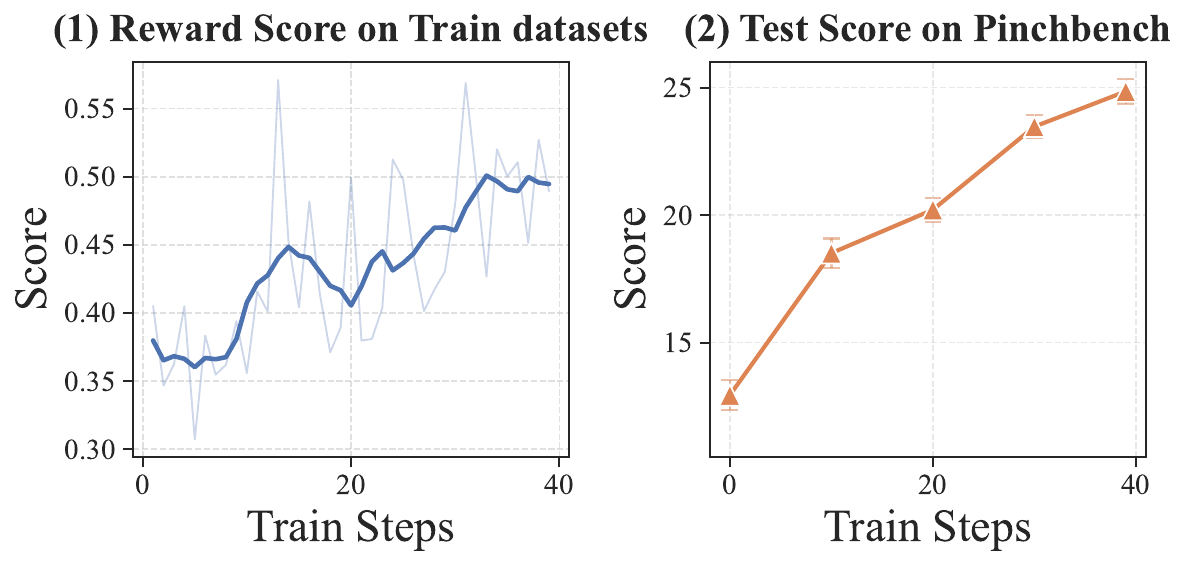}
        \caption{EnvCraft-Claw on Qwen3-8B}
    \end{subfigure}
    \begin{subfigure}[b]{0.46\textwidth}
        \centering
        \includegraphics[width=\linewidth]{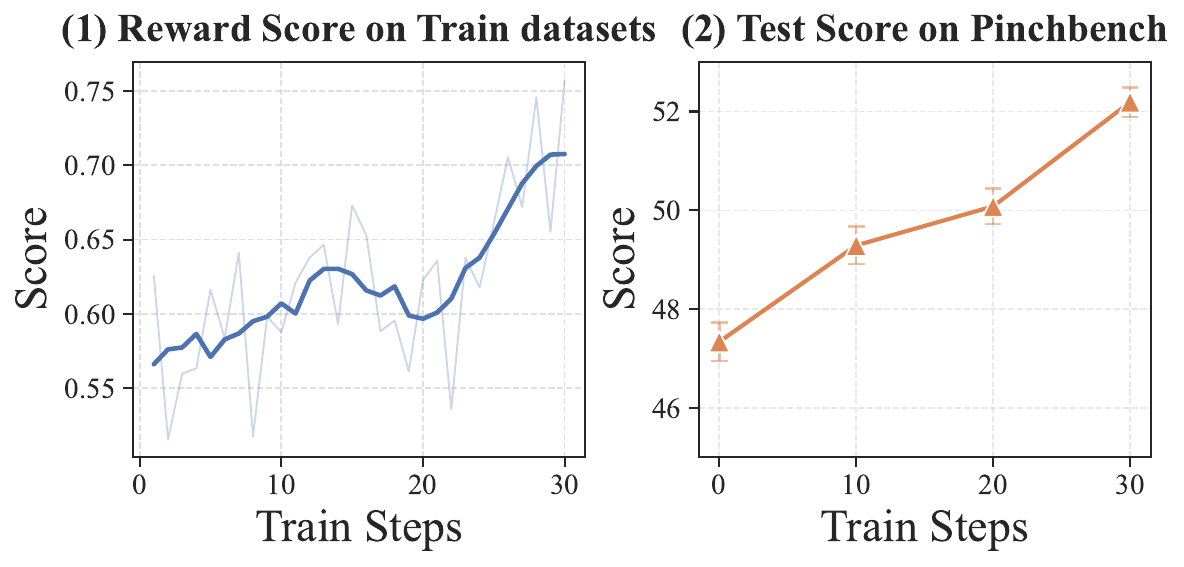}
        \caption{EnvCraft-Claw on Qwen3.5-9B}
    \end{subfigure}

    \caption{
    Training dynamics on EnvCraft-Claw environments. 
    Test PinchBench scores are reported throughout RL, showing consistent and stable improvements for both (a) Qwen3-8B and (b) Qwen3.5-9B as training progresses.
    }
    \label{fig:training_claw}
\end{figure}

\textbf{Training dynamics.} 
Figure~\ref{fig:training_claw} illustrates the training dynamics of Qwen3-8B and Qwen3.5-9B during reinforcement learning on EnvCraft-Claw environments, reporting checkpoint-level test performance on PinchBench. 
Both models exhibit steady improvements throughout training: Qwen3-8B rises from 12.9\% to 24.85\%, while Qwen3.5-9B climbs from 47.3\% to 52.2\%. 
These consistent gains, across different scales and backbones, demonstrate that EnvCraft delivers stable, transferable learning signals that accumulate through sustained interaction with executable environments, rather than arising from isolated checkpoints.

\paragraph{Results on Tool-Use Benchmarks.}
Table~\ref{general_results} presents results on BFCL-v3 and $\tau^2$-bench, evaluating general tool-use capabilities, a core sub-ability underlying claw-style tasks. 
EnvCraft training consistently improves scores across both benchmarks. 
On BFCL, Qwen3-8B climbs from 39.25\% to 46.25\% under EnvCraft-Tool (+7.00) and achieves 45.88\% under EnvCraft-Claw, matching the performance of Kimi-K2-Instruct (45.88\%) despite being substantially smaller. 
Qwen3-32B gains +5.50, reaching 51.00\%, while Qwen3.5-9B shows the largest boost, rising from 44.75\% to 52.75\% (+8.00). 
On $\tau^2$-bench, EnvCraft-Tool improves Qwen3-8B from 27.09\% to 29.80\% (+2.71), while EnvCraft-Claw yields 28.57\%.
Notably, EnvCraft-Claw achieves results on BFCL comparable to EnvCraft-Tool (45.88\% vs. 46.25\% for Qwen3-8B), indicating that Claw-specific training induces transferable tool-use behavior. 

\begin{figure}[t]
    \centering
    \includegraphics[width=0.8\linewidth]{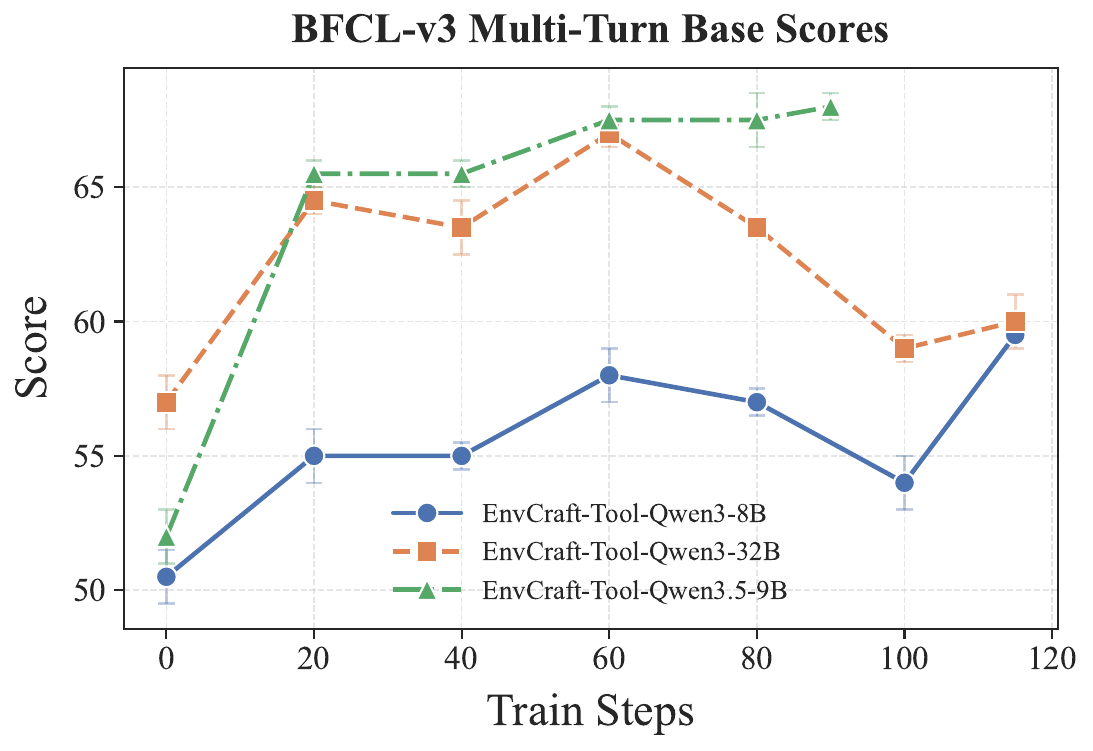}
    \caption{Training dynamics on the BFCL-v3 Base subset.
    We plot the Base score of Qwen3-8B, Qwen3-32B, and Qwen3.5-9B against RL training steps on EnvCraft.
    }
    \label{fig:bfcl_base_comparison}
\end{figure}

\textbf{Training dynamics.}
Figure~\ref{fig:bfcl_base_comparison} tracks the BFCL-v3 Base score of the three Qwen backbones as a function of RL training steps on EnvCraft.
All three models improve over their step-0 initialization within the first $\approx$20 steps and stay above it for the remainder of training, reaching selected-checkpoint gains of 8.5, 10.0, and 13.5 points on Base (Table~\ref{general_results}).
Qwen3.5-9B improves most monotonically, whereas the two Qwen3
backbones exhibit mild mid-training oscillations but remain above
their initial performance.


\begin{figure}[t]
    \centering
    \includegraphics[width=0.8\linewidth]{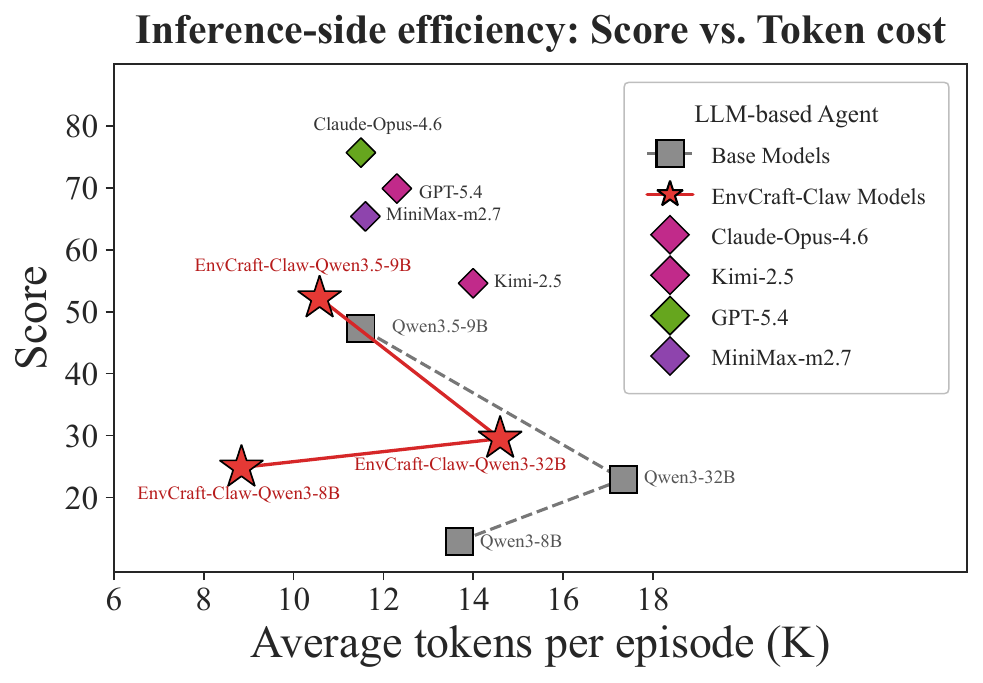}
    \caption{
    Comparison of PinchBench performance and average token consumption per task before and after EnvCraft-Claw training. 
    Higher task scores and lower token costs indicate better inference efficiency.
    }
    \label{fig:inference_efficiency}
\end{figure}

 \paragraph{Inference Token-Cost.}
Figure~\ref{fig:inference_efficiency} compares task performance and average token consumption on PinchBench before and after EnvCraft-Claw training. Across all three backbones, EnvCraft-Claw reduces per-task token cost by an average of 20\% (42.5K $\rightarrow$ 34.0K in aggregate) while simultaneously improving task scores. Specifically, Qwen3-8B drops from 13.69K to 8.84K tokens ($-35\%$), Qwen3-32B from 17.34K to 14.60K ($-16\%$), and Qwen3.5-9B from 11.50K to 10.58K ($-8\%$).
The consistent score-up, cost-down pattern indicates that RL on stateful environments encourages models to discover shorter, more direct action sequences rather than verbose trial-and-error trajectories.

\section{Conclusion}
We present EnvCraft, an end-to-end synthesis framework that automates the generation of executable environments and verifiable RL training tasks from natural-language descriptions, eliminating the need for manual environment engineering. 
We contribute 139 RL environments and $\sim20K$ training instances at scale. 
Experiments demonstrate consistent gains on both Claw-style and general tool-use benchmarks, accompanied by reduced inference token costs.
These results suggest that scalable, automated environment synthesis is a viable way to train claw-like agents.

\bibliography{aaai2026}

\appendix

\input{./appdx}

\end{document}

%% file: appdx.tex
\section*{Limitations}
This paper has several limitations:

First, although the synthesized environments are executable and undergo static and dynamic auditing, they remain simplified abstractions of real-world systems and may not fully capture external-service failures, nondeterministic behavior, evolving interfaces, or complex security constraints encountered in deployment. 

Second, the synthesis pipeline relies on LLM-generated specifications, dependency graphs, user intents, and verification scripts. Consequently, biases or errors introduced by the generator may propagate across different stages, while the predefined interaction archetypes and topology-based sampling strategy may underrepresent workflows outside the covered design patterns. 

Third, our state-based deterministic verifiers primarily evaluate whether the intended workspace mutations are completed, making them less suitable for tasks involving subjective quality, multiple acceptable outcomes, or purely read-oriented objectives. 

Finally, our training experiments focus on Qwen3 and Qwen3.5 backbones and a limited set of agent benchmarks. Evaluating broader model families, real production systems, and longer-running workflows remains an important direction for future work.


\section*{Ethical Statement}
EnvCraft aims to reduce the cost of constructing executable environments and training data for agentic reinforcement learning. 
Nevertheless, the proposed capabilities involve potential risks. 
We therefore emphasize that EnvCraft is intended for controlled research environments. Generated environments and tasks should undergo security auditing, access control, and human review before release or deployment, particularly when they involve sensitive data, external services, or tools capable of producing consequential state changes.

\section*{Generative AI Statement}
LLMs was only used to refine writing. 
Generative AI tools were not used to formulate the scientific hypotheses, interpret experimental results, derive mathematical formulations, or develop the scientific arguments presented in this manuscript. 
They were also not used to draft, or substantively alter the manuscript text. 
The authors take full responsibility for the design of the study, the validity of the experiments, the interpretation of the results, and the accuracy and integrity of the manuscript.

\section{Related Works}
\paragraph{Agentic Reinforcement Learning}

Recent research has increasingly shifted from supervised fine-tuning
(SFT) toward Agentic RL for learning complex tool-use and multi-turn
interaction policies. By enabling autonomous exploration in interactive
environments and optimizing policies from rule-based or trajectory-level
feedback, methods such as GRPO~\citep{guo2025deepseek} and
DAPO~\citep{yu2025dapo} improve model performance on challenging and
out-of-distribution tasks. This closed loop between execution and learning
allows agents to recover from errors and refine their behavior using
environment feedback.

For general tool use, methods such as ToolRL~\citep{qian2026toolrl},
Tool-Zero~\citep{zeng2025tool}, and ToolN1~\citep{zhang2025nemotron}
train LLMs to select and invoke tools by directly optimizing task success.
Search-oriented agents~\citep{li2026webthinker,wu2026webdancer,li2025websailor}
interleave reasoning with real-time evidence retrieval to address
challenging open-domain questions. In mathematical reasoning, ReTool,
ToRL, and SimpleTIR~\citep{feng2025retool,li2025torl,xue2025simpletir}
treat interpreters as executable action spaces. Similarly, CodeRL and
DataMind~\citep{le2022coderl,qiao2025scaling} optimize code-generation
policies using execution feedback. These developments further increase
the demand for diverse, executable, and scalable training environments.

\paragraph{Executable Environment Synthesis}

Large-scale agent RL training demands low-latency, scalable, and
reproducible environments. Environment construction can be broadly
divided into direct production environments and synthesized
environments. The latter can be further categorized into simulated
environments and executable synthetic environments.

\textit{Production environments} employ real-world APIs
\citep{qin2024toolllm} and MCP servers \citep{xu2025toucan} to provide
authentic execution. However, they are expensive to scale and suffer
from network latency and irreproducible states, which may destabilize
RL training.

\textit{Simulated environments} leverage LLMs to emulate tool behavior
and state dynamics, enabling rapid prototyping
\citep{chen2025scaling,li2025simulating,li2026word}. However, they are
prone to hallucination and introduce additional inference cost and
instability, limiting their applicability to large-scale training
\citep{kalai2025language,wang2024can}.

\textit{Executable synthetic environments}
\citep{wang2026agent,song2026envscaler,xu2026envfactory,
tu2026scaleenv,fang2026towards}
reconstruct tools and stateful backends through sandbox code generation,
striking a balance among scalability, reproducibility, and execution
fidelity.

\section{Additional Method Details}
\subsection{Group Relative Policy Optimization}
\label{sec:grpo}

We briefly review GRPO~\citep{shao2024deepseekmath} in the language-generation setting.
Formally, for each input instruction $x$, the policy $\pi_\theta$ samples a group of $G$ candidate responses $\{y_i\}_{i=1}^G$. 
The optimization objective is defined as:
\begin{equation}
\bar{\rho}_{i,t}
=
\operatorname{clip}
\left(
\rho_{i,t},
1-\epsilon_{\mathrm{clip}},
1+\epsilon_{\mathrm{clip}}
\right).
\end{equation}

\begin{equation}
\small
\begin{aligned}
\mathcal{J}(\theta)
=
\mathbb{E}\Bigg[
\frac{1}{G}
\sum_{i=1}^{G}
\frac{1}{|y_i|}
\sum_{t=1}^{|y_i|}
\Big(
\min\{
\rho_{i,t}\hat{A}_i,
\bar{\rho}_{i,t}\hat{A}_i
\}
-
\beta D_{\mathrm{KL},i,t}
\Big)
\Bigg].
\end{aligned}
\end{equation}
Here, $y_i$ denotes the $i$-th sampled agent trajectory, and
$r_i$ is the outcome reward returned by the deterministic task verifier.
Because the reward is assigned at the trajectory level, we use
$\hat{A}_{i,t}=\hat{A}_i$ for all tokens in trajectory $i$.

\subsection{Environment Scenario Taxonomy}
\label{sec:environment_scenario_taxonomy}

We organize the collected environment domains into 83 fine-grained
scenarios, including 41 general tool-use scenarios and 42
Claw-specific scenarios. Each fine-grained scenario may correspond to
one or more executable environment instances. The ``\# Envs'' column
reports the number of synthesized environments associated with each
scenario. Tables~\ref{tab:general_tool_scenarios} and
\ref{tab:claw_specific_scenarios} provide the complete taxonomy and
scenario descriptions.


\begin{table*}[t]
\centering
\footnotesize
\setlength{\tabcolsep}{3pt}
\renewcommand{\arraystretch}{1.12}
\begin{tabular}{
    @{}
    C{0.035\textwidth}
    L{0.245\textwidth}
    C{0.055\textwidth}
    L{0.585\textwidth}
    @{}
}
\toprule
\textbf{No.} &
\textbf{Fine-Grained Scenario} &
\textbf{\# Envs} &
\textbf{Scenario Description} \\
\midrule

1 &
Messaging \& Communication &
5 &
APIs for instant messaging, mobile messaging, email, SMS, and other
communication services, supporting session management, message
sending and receiving, and user-status tracking. \\

2 &
Document \& Note Management &
3 &
APIs for document-management and personal note-taking systems,
supporting document creation and editing, version control, note
organization, and information retrieval. \\

3 &
Development Tools &
1 &
Jupyter Notebook server APIs supporting the creation, execution,
editing, and lifecycle management of computational notebooks. \\

4 &
Scheduling \& Reminders &
2 &
APIs for scheduling and reminder systems, supporting calendar-event
management, timed reminders, recurrence configuration, and
notification delivery. \\

5 &
Monitoring \& Logging &
3 &
APIs for network monitoring, model monitoring, and system-log
management, supporting device-status detection, log collection,
alerting, and performance-metric tracking. \\

6 &
Server \& Application Management &
1 &
Node.js application-server management APIs supporting application
deployment, start and stop operations, process management, and runtime
status monitoring. \\

7 &
Data Management &
3 &
APIs for database management, data backup and recovery, and file-system
operations, supporting persistent storage, data restoration, and
structured file manipulation. \\

8 &
Memory \& Storage &
2 &
Key--value storage and memory-summarization APIs providing persistent
information storage, retrieval, summarization, and archival-memory
management. \\

9 &
Air Travel &
1 &
Airline-ticketing APIs supporting flight search, seat reservation,
payment processing, itinerary retrieval, and booking modification. \\

\bottomrule
\end{tabular}
\caption{
Fine-grained general tool-use scenarios and their corresponding
environment descriptions.
}
\label{tab:general_tool_scenarios}
\end{table*}

\begin{table*}[t]
\ContinuedFloat
\centering
\footnotesize
\setlength{\tabcolsep}{3pt}
\renewcommand{\arraystretch}{1.12}
\begin{tabular}{
    @{}
    C{0.035\textwidth}
    L{0.245\textwidth}
    C{0.055\textwidth}
    L{0.585\textwidth}
    @{}
}
\toprule
\textbf{No.} &
\textbf{Fine-Grained Scenario} &
\textbf{\# Envs} &
\textbf{Scenario Description} \\
\midrule

10 &
Ground Transport &
2 &
Car-rental and ride-hailing platform APIs supporting vehicle leasing,
ride requests, driver and vehicle matching, route planning, and fare
calculation. \\

11 &
Comprehensive Travel Booking &
1 &
Integrated travel-booking APIs combining flight reservations,
credit-card management, payment processing, and travel-budget
constraints. \\

12 &
Accommodation &
1 &
Hotel-booking APIs supporting hotel and room search, availability
management, price calculation, booking creation, modification, and
cancellation. \\

13 &
Vehicle Management &
3 &
APIs for vehicle maintenance, motor-vehicle registration, and vehicle
control, covering maintenance scheduling, registration records, and
in-vehicle system operations. \\

14 &
Banking \& Trading &
3 &
APIs for retail banking, stock trading, and quantitative-trading bots,
supporting account management, transfers, market-data retrieval, order
placement, and automated trading. \\

15 &
Corporate Finance &
1 &
Corporate financial-reporting APIs supporting financial-statement
generation, multi-source data aggregation, accounting-period
management, and compliance reporting. \\

16 &
Insurance &
2 &
APIs for insurance-policy management and automobile-insurance claims,
supporting policy lifecycle management, claim submission, evidence
review, and settlement processing. \\

17 &
Reimbursement &
1 &
Reimbursement and expense-claim management APIs supporting application
submission, approval workflows, expense verification, and financial
write-off processing. \\

\bottomrule
\end{tabular}
\caption{
Fine-grained general tool-use scenarios and their corresponding
environment descriptions (continued).
}
\end{table*}

\begin{table*}[t]
\ContinuedFloat
\centering
\footnotesize
\setlength{\tabcolsep}{3pt}
\renewcommand{\arraystretch}{1.12}
\begin{tabular}{
    @{}
    C{0.035\textwidth}
    L{0.245\textwidth}
    C{0.055\textwidth}
    L{0.585\textwidth}
    @{}
}
\toprule
\textbf{No.} &
\textbf{Fine-Grained Scenario} &
\textbf{\# Envs} &
\textbf{Scenario Description} \\
\midrule

18 &
Online Shopping \& Cart &
2 &
APIs for online-shopping and retail platforms, supporting product
browsing, cart management, promotional discounts, checkout, and order
creation. \\

19 &
Product Catalog Management &
1 &
E-commerce product-catalog APIs supporting product-information
maintenance, category organization, attribute management, pricing, and
inventory display. \\

20 &
Inventory \& Order Management &
2 &
APIs for inventory and order management, supporting inventory updates,
order-status transitions, fulfillment tracking, and stock
reconciliation. \\

21 &
Marketing \& Reviews &
2 &
APIs for product promotion and review systems, supporting campaign
configuration, promotional-rule management, user reviews, ratings, and
feedback moderation. \\

22 &
Task \& Issue Tracking &
3 &
APIs for task tracking, to-do management, and software-testing
workflows, supporting assignment, prioritization, status transitions,
and defect tracking. \\

23 &
Workflow \& Job Scheduling &
2 &
APIs for workflow-management and job-scheduling systems, supporting
process definition, dependency configuration, job execution, retry
handling, and status monitoring. \\

24 &
Content Management &
1 &
Web content-management system APIs supporting page creation, editing,
version control, approval, publication, and content lifecycle
management. \\

25 &
Media Catalog \& Playback &
2 &
APIs for media-catalog databases and music-playback systems, supporting
media indexing, metadata management, streaming playback, and playlist
organization. \\

\bottomrule
\end{tabular}
\caption{
Fine-grained general tool-use scenarios and their corresponding
environment descriptions (continued).
}
\end{table*}

\begin{table*}[t]
\ContinuedFloat
\centering
\footnotesize
\setlength{\tabcolsep}{3pt}
\renewcommand{\arraystretch}{1.12}
\begin{tabular}{
    @{}
    C{0.035\textwidth}
    L{0.245\textwidth}
    C{0.055\textwidth}
    L{0.585\textwidth}
    @{}
}
\toprule
\textbf{No.} &
\textbf{Fine-Grained Scenario} &
\textbf{\# Envs} &
\textbf{Scenario Description} \\
\midrule

26 &
Social Media &
1 &
Social-media publishing APIs supporting post creation, comment
interaction, reposting, content retrieval, and follower-relationship
management. \\

27 &
HR \& Employee Management &
1 &
Human-resources information-system APIs supporting employee-profile
management, organizational structures, employment records, and payroll
information maintenance. \\

28 &
Account \& Membership &
2 &
APIs for membership management and user registration, supporting
account creation, authentication, membership-tier management, and
access-control configuration. \\

29 &
Benefits \& Wellness &
1 &
Fitness-class booking APIs supporting class scheduling, member
registration, coach assignment, capacity management, and booking
cancellation. \\

30 &
Communication \& Services &
2 &
APIs for telecommunications and food-delivery services, supporting
mobile-plan management, data-usage control, restaurant ordering, and
delivery-status tracking. \\

31 &
Search \& Information &
1 &
Web-search APIs supporting keyword queries, result retrieval, result
organization, search-history management, and bookmark collection. \\

32 &
Computation \& Math &
1 &
Mathematical-computation APIs supporting logarithmic, trigonometric,
statistical, and numerical functions, together with configurable
precision control. \\

33 &
Online Learning Platforms &
1 &
Online-education platform APIs supporting course management, student
enrollment, progress tracking, assignment handling, and Q\&A
interactions. \\

\bottomrule
\end{tabular}
\caption{
Fine-grained general tool-use scenarios and their corresponding
environment descriptions (continued).
}
\end{table*}

\begin{table*}[t]
\ContinuedFloat
\centering
\footnotesize
\setlength{\tabcolsep}{3pt}
\renewcommand{\arraystretch}{1.12}
\begin{tabular}{
    @{}
    C{0.035\textwidth}
    L{0.245\textwidth}
    C{0.055\textwidth}
    L{0.585\textwidth}
    @{}
}
\toprule
\textbf{No.} &
\textbf{Fine-Grained Scenario} &
\textbf{\# Envs} &
\textbf{Scenario Description} \\
\midrule

34 &
Academic Administration &
3 &
APIs for student-information and university course-selection systems,
supporting student-profile management, course registration, enrollment
constraints, and grade-record maintenance. \\

35 &
Knowledge Resources &
1 &
Library-management APIs supporting catalog search, book borrowing,
reservations, renewals, return processing, and overdue-fine
calculation. \\

36 &
CRM \& Customer Management &
1 &
Customer-relationship management APIs supporting customer profiles,
sales-opportunity tracking, interaction histories, and communication
record maintenance. \\

37 &
Reservation \& Booking &
1 &
Restaurant-reservation APIs supporting restaurant and table
information management, availability queries, booking creation, and
dining-schedule modification. \\

38 &
Support Ticketing &
1 &
Customer-support ticketing APIs supporting ticket creation, assignment,
status updates, priority management, escalation, and service-queue
management. \\

39 &
Health Records &
1 &
Personal-health-record APIs supporting health-metric recording,
vital-sign tracking, historical-data retrieval, and medical-record
management. \\

40 &
Nutrition \& Fitness &
1 &
Nutrition-tracking and fitness-management APIs supporting dietary
records, nutritional analysis, exercise logging, and workout-plan
management. \\

41 &
Sensor \& Device Management &
1 &
IoT sensor-data platform APIs supporting device registration,
sensor-data collection, device-status monitoring, and time-series data
management. \\

\bottomrule
\end{tabular}
\caption{
Fine-grained general tool-use scenarios and their corresponding
environment descriptions (continued).
}
\end{table*}

\begin{table*}[t]
\centering
\footnotesize
\setlength{\tabcolsep}{3pt}
\renewcommand{\arraystretch}{1.12}
\begin{tabular}{
    @{}
    C{0.035\textwidth}
    L{0.245\textwidth}
    C{0.055\textwidth}
    L{0.585\textwidth}
    @{}
}
\toprule
\textbf{No.} &
\textbf{Fine-Grained Scenario} &
\textbf{\# Envs} &
\textbf{Scenario Description} \\
\midrule

1 &
Collaborative Office \& Email Automation &
3 &
Collaborative-office and email-automation environments supporting
workflow orchestration for email-content extraction, automated
social-media publishing, and event-driven email rules. \\

2 &
E-commerce \& Logistics Management &
5 &
E-commerce logistics environments supporting shipment tracking,
return and exchange processing, inventory reconciliation, multi-stage
logistics orchestration, and state-machine-based order management. \\

3 &
Spreadsheet Data Cleaning \& Statistical Analysis &
1 &
Spreadsheet-analysis environments supporting data cleaning, formula
calculation, chart generation, pivot-table construction, and
statistical analysis across workspace files. \\

4 &
Email Client Intelligent Management &
1 &
Intelligent email-client environments supporting email sending and
receiving, message organization, reply suggestion, and automatic
extraction of actionable to-do items. \\

5 &
Sensitive Information Encryption \& Management &
1 &
Secure information-storage environments supporting encrypted password
storage, credential retrieval, automatic form filling, and strong
password generation. \\

6 &
Cross-Modal Content Transcription \& Summarization &
2 &
Cross-modal content-processing environments supporting audio- and
video-to-text transcription, discussion consolidation, and
multi-turn conversational summarization. \\

7 &
Personal Assistant Configuration &
2 &
Personal-assistant environments supporting assistant-persona
configuration, preference management, schedule coordination, and
multi-turn dialogue orchestration. \\

8 &
Contact Information Intelligent Management \& Maintenance &
1 &
Intelligent contact-management environments supporting contact
storage, tag-based classification, relationship maintenance, and
context-aware reminders. \\

9 &
Travel Compliance \& Booking Closure &
5 &
Travel-compliance environments supporting policy verification,
multi-platform price comparison, approval workflows, booking
completion, and compliance-oriented state transitions. \\

\bottomrule
\end{tabular}
\caption{
Fine-grained Claw-specific scenarios and their corresponding
environment descriptions.
}
\label{tab:claw_specific_scenarios}
\end{table*}

\begin{table*}[t]
\ContinuedFloat
\centering
\footnotesize
\setlength{\tabcolsep}{3pt}
\renewcommand{\arraystretch}{1.12}
\begin{tabular}{
    @{}
    C{0.035\textwidth}
    L{0.245\textwidth}
    C{0.055\textwidth}
    L{0.585\textwidth}
    @{}
}
\toprule
\textbf{No.} &
\textbf{Fine-Grained Scenario} &
\textbf{\# Envs} &
\textbf{Scenario Description} \\
\midrule

10 &
Itinerary Disruption Dynamic Optimization &
4 &
Dynamic itinerary-optimization environments supporting delay
monitoring, automatic rebooking, linked hotel and airport-transfer
adjustments, and traveler notification. \\

11 &
Travel Budget Auto-Accounting \& Control &
1 &
Automated travel-budget environments supporting expense
categorization, budget allocation, threshold alerts, and spending-trend
analysis. \\

12 &
Automated Reimbursement Reconciliation &
5 &
Automated reimbursement environments supporting receipt recognition,
bank-statement matching, expense-report completion, approval
processing, and multidimensional reconciliation. \\

13 &
Travel Itinerary Offline Planning &
1 &
Offline travel-planning environments supporting destination
organization, itinerary construction, local information management,
and multi-objective route optimization. \\

14 &
Multi-Dimensional Environmental Adaptive Regulation &
3 &
Adaptive environmental-control environments supporting smart-device
management, HVAC regulation, air-quality adjustment, and
energy-consumption optimization. \\

15 &
Home/Office Security Patrol &
4 &
Security-patrol environments supporting intrusion detection, scheduled
inspection, alarm triggering, coordinated incident response, and
evidence collection. \\

16 &
Environmental Data Real-Time Monitoring \& Statistics &
1 &
Real-time environmental-monitoring environments supporting sensor-data
collection, threshold alerts, temporal trend analysis, and statistical
report generation. \\

17 &
Smart Device Scheduled Task Management &
1 &
Smart-device scheduling environments supporting access control, area
monitoring, scheduled patrol tasks, rule execution, and emergency
notification. \\

18 &
Investment Research Cross-Validation &
4 &
Investment-research environments supporting financial-data comparison,
stock-price verification, research-report generation, and real-time
market-stream processing. \\

\bottomrule
\end{tabular}
\caption{
Fine-grained Claw-specific scenarios and their corresponding
environment descriptions (continued).
}
\end{table*}

\begin{table*}[t]
\ContinuedFloat
\centering
\footnotesize
\setlength{\tabcolsep}{3pt}
\renewcommand{\arraystretch}{1.12}
\begin{tabular}{
    @{}
    C{0.035\textwidth}
    L{0.245\textwidth}
    C{0.055\textwidth}
    L{0.585\textwidth}
    @{}
}
\toprule
\textbf{No.} &
\textbf{Fine-Grained Scenario} &
\textbf{\# Envs} &
\textbf{Scenario Description} \\
\midrule

19 &
Industry Sentiment \& Growth Analysis &
1 &
Industry-analysis environments supporting competitor monitoring,
strategy tracking, market-report generation, growth analysis, and
user-feedback aggregation. \\

20 &
System Patch Injection Defense &
1 &
System-security environments supporting sensitive-key isolation,
prompt-injection detection, suspicious patch inspection, and
security-audit alerting. \\

21 &
Identity Forgery \& Audit Fraud &
1 &
Identity-verification and audit-defense environments supporting
auditor-identity validation, anomalous-behavior detection, forgery-risk
assessment, and early warning. \\

22 &
Bulk Unauthorized Data Export Interception &
1 &
Data-loss-prevention environments supporting sensitive-data
identification, export-approval workflows, permission verification,
and unauthorized-transfer blocking. \\

23 &
Internal Core Document Leak Interception &
1 &
Document-security environments supporting confidentiality
classification, outbound-transfer detection, access-policy validation,
and real-time leak prevention. \\

24 &
End-to-End Interview Process Hosting &
1 &
Recruitment-workflow environments supporting resume screening,
interview scheduling, candidate communication, evaluator coordination,
and end-to-end status tracking. \\

25 &
Employee Onboarding Automated Pipeline &
1 &
Employee-onboarding environments supporting account creation, asset
allocation, permission configuration, document processing, and
automatic training-task triggering. \\

26 &
Employee Performance Auto-Assessment Archiving &
1 &
Performance-management environments supporting multi-source employee
data aggregation, score calculation, report generation, and persistent
assessment archiving. \\

\bottomrule
\end{tabular}
\caption{
Fine-grained Claw-specific scenarios and their corresponding
environment descriptions (continued).
}
\end{table*}

\begin{table*}[t]
\ContinuedFloat
\centering
\footnotesize
\setlength{\tabcolsep}{3pt}
\renewcommand{\arraystretch}{1.12}
\begin{tabular}{
    @{}
    C{0.035\textwidth}
    L{0.245\textwidth}
    C{0.055\textwidth}
    L{0.585\textwidth}
    @{}
}
\toprule
\textbf{No.} &
\textbf{Fine-Grained Scenario} &
\textbf{\# Envs} &
\textbf{Scenario Description} \\
\midrule

27 &
Employee Offboarding Automated Process &
1 &
Employee-offboarding environments supporting permission revocation,
asset-return confirmation, workspace-data backup, account
deactivation, and completion tracking. \\

28 &
Automated Domain Technical Survey &
1 &
Automated technical-survey environments supporting paper retrieval,
source organization, domain-trend analysis, evidence synthesis, and
survey-report generation. \\

29 &
Literature Citation Relationship Archiving &
1 &
Literature-management environments supporting citation-graph
construction, citation-impact analysis, related-work discovery, and
persistent relationship archiving. \\

30 &
Multi-Group Experiment Metrics Auto-Comparison &
1 &
Experiment-analysis environments supporting experimental-configuration
recording, metric aggregation, group comparison, statistical
difference analysis, and result archiving. \\

31 &
Open-Source Reproduction Record Archiving &
1 &
Reproduction-management environments supporting procedural record
keeping, dependency and environment configuration, result validation,
and reproducibility archiving. \\

32 &
Music \& Audio Playback &
1 &
In-vehicle audio environments supporting music search, playback
control, queue manipulation, playlist management, and media-status
retrieval. \\

33 &
Vehicle Control \& Status Management &
1 &
Vehicle-control environments supporting air-conditioning, seat,
window, lighting, driving-mode, and multimedia operations together
with vehicle-status inspection. \\

34 &
Navigation \& Trip Planning &
1 &
In-vehicle navigation environments supporting destination search,
route planning, real-time traffic retrieval, charging-station
selection, and trip adjustment. \\

\bottomrule
\end{tabular}
\caption{
Fine-grained Claw-specific scenarios and their corresponding
environment descriptions (continued).
}
\end{table*}

\begin{table*}[t]
\ContinuedFloat
\centering
\footnotesize
\setlength{\tabcolsep}{3pt}
\renewcommand{\arraystretch}{1.12}
\begin{tabular}{
    @{}
    C{0.035\textwidth}
    L{0.245\textwidth}
    C{0.055\textwidth}
    L{0.585\textwidth}
    @{}
}
\toprule
\textbf{No.} &
\textbf{Fine-Grained Scenario} &
\textbf{\# Envs} &
\textbf{Scenario Description} \\
\midrule

35 &
Infrastructure Automated Operations &
1 &
Infrastructure-operations environments supporting server-fault
detection, diagnostic inspection, automatic recovery, service
verification, and operations-record maintenance. \\

36 &
Cloud Resource Cost Auto-Audit &
1 &
Cloud-cost auditing environments supporting multi-project cost
aggregation, abnormal-spending detection, budget alerts, and
cost-optimization recommendations. \\

37 &
Online Fault Root-Cause Postmortem Archiving &
1 &
Incident-management environments supporting fault-timeline
construction, root-cause analysis, corrective-action tracking, and
postmortem knowledge-base archiving. \\

38 &
Brand \& Product Research &
1 &
Brand- and product-research environments supporting SKU-information
retrieval, competitor-feature comparison, evidence aggregation, and
market-positioning analysis. \\

39 &
Multimedia Clue Tracking &
1 &
Multimedia-investigation environments supporting cross-document clue
association, evidence-chain construction, provenance tracking, and
information tracing. \\

40 &
Churn Warning \& Proactive Retention &
1 &
Customer-retention environments supporting churn-risk assessment,
retention-strategy generation, customer segmentation, and automated
customer communication. \\

41 &
Customer Tier Label Auto-Classification &
1 &
Customer-classification environments supporting behavior analysis,
tier-rule configuration, eligibility evaluation, and automatic
customer-label updates. \\

42 &
Scheduled Business Report Auto-Generation &
1 &
Scheduled reporting environments supporting Markdown-template
completion, multi-source data aggregation, visualization generation,
and scheduled report publication. \\

\bottomrule
\end{tabular}
\caption{
Fine-grained Claw-specific scenarios and their corresponding
environment descriptions (continued).
}
\end{table*}
\subsection{Interaction Archetype Specifications}
To systematically address the complex operational demands inherent in OpenClaw scenarios, 
such as multi-tool orchestration, asynchronous agentic execution, and dynamic context shifts,
we define a taxonomy of nine complementary interaction archetypes. 
Derived from core software interaction patterns in real-world OpenClaw deployments, these prototypes encapsulate the exhaustive spectrum of control flows and state evolution logic required for agent evaluation.

\begin{enumerate}
    \item \textbf{Transactional Workspace Operations (CRUD \& Auth):}
    Models authenticated entity workspaces---ticketing systems, content platforms, multi-tenant
    SaaS---where a session-bound user handle gates operations on a typed entity queue, each entity
    carrying an immutable identifier, mutable payload, and discrete status marker that progresses
    unidirectionally (open $\to$ in-progress $\to$ resolved $\to$ closed) with guard checks against
    illegal reversions and duplicate terminal closure; an append-only audit log records every
    mutation. Tools expose entity CRUD with schema validation, selective field updates,
    terminal-state resolution with mandatory metadata, session lifecycle primitives, and user-scoped
    filtered listing. The reference blueprint is a stateful workspace class: a module-level
    \texttt{DEFAULT\_STATE} dict holds the entity list, identity counter, current-user variable, and
    log; lifecycle methods validate authentication and schema constraints before mutating; a
    \texttt{\_load\_scenario} entry point deep-copies defaults and applies overrides for harness
    snapshot-consistency.

    \item \textbf{Cross-Tool Data Pipelines:}
    Simulates multi-stage ETL workflows---media monitoring, competitive intelligence, multi-source
    research synthesis---where heterogeneous raw inputs (text queries, audio, video, structured
    tables) are ingested through format-specific parsers into a unified intermediate representation,
    each job tracked with an auto-derived type tag (search, transcribe, parse) and processing
    status; completed jobs feed an aggregation stage that merges results with semantic deduplication
    across configurable keys, and a final rendering stage formats the consolidated set into markdown,
    bullet summaries, JSON, or graph descriptions. Tools provide source registration, multi-format
    ingestion, type-dispatched processing, cross-job aggregation with dedup-key configuration, and
    multi-format output generation. The reference architecture is a four-phase pipeline engine
    (ingest $\to$ process $\to$ aggregate $\to$ render) where each phase produces an intermediate
    artifact; processing uses a dispatch table keyed by job-type to select among deterministic
    simulator functions; aggregation operates on a user-declared deduplication key with configurable
    string-normalization; rendering follows a format-strategy pattern.

    \item \textbf{Dynamic Human-Agent Interaction:}
    Replicates multi-turn service conversations---customer-support triage, consultative sales,
    medical intake---where the agent must discover a hidden user persona (latent intent,
    floating-point confidence level, emotional valence and trend, confirmed and outstanding item
    sets) through progressive clarification, with a patience threshold forcing convergence or
    escalation when exceeded. State is organized into three layers: session records with turn
    counts, per-session message histories, and context-variable dictionaries; control flow runs a
    keyword-driven analysis pipeline that adjusts intent confidence (raised by confirmation
    language, lowered by clarification), transfers items from outstanding to confirmed on resolution
    detection, tracks mood deterioration beyond the patience bound, and generates deterministic
    mood-conditioned replies from a template bank seeded by message content and turn index. Tools
    expose session start with persona configuration, message send with auto-reply and context-delta
    reporting, history retrieval, manual context overrides, and session summarization with
    trajectory analysis. The blueprint is a three-layer state container with a pure analysis
    function mapping message text to context deltas and a deterministic template-based reply
    generator enabling reproducible evaluation.

    \item \textbf{Asynchronous Event Triggering:}
    Captures event-condition-action automation---IFTTT/Zapier-style hubs, monitoring pipelines,
    smart-environment rule engines---where agents register heterogeneous event sources from a fixed
    taxonomy of external system categories, define rules binding a trigger type, optional source
    filter, and a condition predicate drawn from an operator set (equality, comparison, containment,
    regex matching) to typed action lists. Event injection is the sole triggering primitive: the
    event is validated, appended to an event log, then synchronously matched against all enabled
    rules with aligned trigger type and source filter; for each match, the condition is evaluated
    against the event payload and, when satisfied, all actions dispatch in definition order with
    results written to an action log. Tools cover source registration, rule CRUD with full
    condition/action specification, event injection, and separate filtered queries over event and
    action histories for end-to-end traceability. The reference architecture is a synchronous ECA
    engine: rule evaluation is a linear scan performed immediately on injection; condition
    evaluation is a pure function with well-defined fallback semantics (string ops degrade
    gracefully, numeric ops fail closed); action execution is a log-only side effect; the dual-log
    design ensures full auditability.

    \item \textbf{Constraint-Guided State Machine:}
    Represents systems governed by an explicitly declared finite state machine $\mathcal{S} \times
    \mathcal{A} \rightarrow \mathcal{S}'$---order lifecycles, loan origination, regulatory approval
    chains, device controllers---where machine definitions enumerate named states with
    on-enter/on-exit hooks, designated initial and terminal sets, and a transition table whose edges
    carry source, target, trigger-mode label, and optional nested guard predicates of the form
    \texttt{\{field: \{op: value\}\}} evaluated with multi-field AND semantics against per-entity
    data payloads. Entities are created at the initial state and transitioned by locating matching
    edges to the target, evaluating guards (fail-closed on missing fields), and on success updating
    current state, appending to visitation history, and logging; terminal-state entry marks the
    entity completed and blocks further transitions. Tools expose declarative machine definition
    (create, add states, add guarded transitions), entity lifecycle with three-result transition
    feedback (transitioned, blocked by guard, no such transition) and available-target enumeration,
    and filtered entity/transition inspection. The blueprint is a two-tier declarative engine: the
    machine tier is a pure graph definition with labeled edges; the entity tier is a runtime
    interpreter that walks the graph per entity; guard validation is a standalone pure function with
    explicit AND composition; trigger mode is metadata for agent interpretation.

    \item \textbf{Multi-Resource Temporal Scheduling:}
    Simulates competitive time-axis resource allocation---calendar management, meeting-room booking,
    shift rostering---where agents manage multiple calendars each with timezone and working-hour
    constraints, and schedule events with ISO-8601 time bounds and integer priority into a global
    event list. Every addition triggers eager pairwise interval-overlap detection against
    non-cancelled events on the same calendar, marking conflicts automatically; resolution follows a
    four-strategy taxonomy: reschedule (assign new time bounds), cancel (terminal status), override
    (cancel all conflicting events of strictly lower priority then promote), or split (truncate at a
    split point and spawn a continuation inheriting all metadata). Availability queries scan a
    target date within working hours, compute gaps via a cursor-based algorithm over sorted
    non-cancelled events, and return contiguous free intervals meeting a minimum duration. Tools
    cover calendar creation, event scheduling with auto-detection, explicit conflict inspection,
    four-strategy resolution, filtered event listing by calendar/status/date prefix, and
    availability discovery. The reference architecture is a temporal-constraint engine: conflict
    detection is a pure interval-overlap predicate; the resolution taxonomy is a strategy pattern
    with cascading override; timestamps use ISO-8601 string comparison for ordering with
    \texttt{datetime} parsing only for duration arithmetic.

    \item \textbf{Multi-Criteria Bilateral Matching:}
    Evaluates two-sided market decision-making---resume-to-job matching, supplier sourcing, roommate
    pairing, marketplace recommendation---where agents configure matching pools with weighted
    criteria vectors (attribute-name to float-weight mappings) and a top-K cap, then populate supply
    and demand candidates with typed attribute dictionaries. Pairwise compatibility scoring
    normalizes per-attribute distances by type---numerical proximity via absolute distance, string
    equality with a partial-match fallback constant, list overlap as Jaccard ratio---and computes a
    weighted sum normalized to 0--100; results are sorted descending, truncated to top-K, and
    materialized as persistent match records that enter a two-phase lifecycle (pending $\to$
    accepted or rejected) with explicit agent-mediated acceptance/rejection on behalf of the
    counterparty. Tools expose pool creation with weighted criteria, candidate addition, pairwise
    match execution with ranked scored output, re-ranking by score/supply/demand, accept/reject
    primitives, and filtered listing. The blueprint is a configurable scoring engine: the criteria
    vector is user-declared; scoring is a type-dispatched per-dimension calculator composed into a
    weighted-sum aggregator; matches are persistent records with an explicit status state machine;
    pools isolate independent matching contexts.

    \item \textbf{Distributed Workflow Orchestration:}
    Replicates complex task-DAG execution---CI/CD pipelines, order fulfillment, content moderation,
    multi-service saga orchestration---where pipelines decompose work into ordered stage arrays, each
    stage specifying an action label, a dependency list of prerequisite stage identities, an assigned
    worker (with role and capability tags), and a failure-policy tag (abort, skip, retry) with a
    retry cap. Execution is mode-dependent: sequential mode iterates stages in order, executing each
    pending stage whose dependencies are satisfied and halting on abort-policy failures; parallel
    mode identifies all dependency-satisfied pending stages and executes them concurrently. On
    failure the policy strategy applies: abort marks the pipeline failed; skip marks the stage
    skipped and continues; retry resets to pending with counter increment for explicit re-invocation.
    Rollback walks backward through completed stages invoking compensating actions; in-band worker
    messaging enables coordination. Tools cover worker registration/deregistration with busy-state
    protection, pipeline definition with sequential/parallel topology, stage-to-worker assignment,
    single-stage and full-pipeline execution, retry, rollback with compensation, result collection,
    and inter-worker messaging. The blueprint is a two-layer orchestration engine: the pipeline layer
    defines DAG topology through dependency lists; the execution layer enforces the dependency
    contract, manages worker state transitions (idle $\leftrightarrow$ busy), and delegates failure
    response to a policy-strategy pattern; rollback is a reverse-topological walk.

    \item \textbf{Fault-Tolerant Stream Processing:}
    Simulates real-time asynchronous publish-subscribe message buses---telemetry ingestion, audit-log
    processing, IoT sensor hubs, event-sourcing backends---where typed event streams are declared
    with explicit schema contracts (field-name-to-type dictionaries), and subscribers attach through
    filtered subscriptions specifying a delivery mode (push, poll, batch) and an exact-equality
    filter condition over event payload fields. Publication pushes a typed payload with an assigned
    priority tier into the stream's append-only history, then fans out to all active matching
    subscriptions, placing event identifiers into per-subscription pending queues; explicit
    acknowledgment removes events from the pending queue, with unacknowledged events persisting
    indefinitely to simulate at-least-once redelivery semantics and create observable backpressure
    through queue depth. Unlike the event-driven pattern (synchronous single-fire rule evaluation),
    stream processing is inherently asynchronous and continuous: events flow indefinitely,
    subscriptions have lifecycles (create, filter, unsubscribe), and delivery guarantees require
    active consumer participation. Tools expose typed stream creation with schema contracts,
    subscription management with filter and delivery-mode selection, prioritized event publication,
    acknowledgment, pending-event inspection, and event-history queries. The reference architecture
    is an asynchronous pub-sub bus: streams are schema-declared channels; subscriptions are
    independent consumer handles each with a private pending queue; the filter-matching function is a
    pure exact-equality predicate; acknowledgment is the sole offset-advancement mechanism.
\end{enumerate}


\section{System Prompts and Execution Schemas}
\label{sec:prompts_schemas}

This section provides the system prompts and structured schemas used by
EnvCraft for environment and task synthesis. To reduce inconsistencies
between natural-language requirements and generated code, both synthesis
stages follow explicit input--output contracts. The environment generator
produces a standalone, session-isolated Python package, while the task
generator constructs a self-contained workspace, a natural user request,
and a deterministic verifier from a shared latent ground truth.

\subsection{Environment Generation Prompt}
\label{sec:environment_generation_prompt}

Given an environment name, a workflow description, a set of domains, and
structured world data, the environment generator is instructed to produce
a complete Python package under
\texttt{claw\_envs/<environment\_name>/}. The package must implement
persistent session management, domain-specific tools, deterministic state
transitions, executable evaluation logic, documentation, and concurrency
tests.

The fixed system prompt used for environment generation is shown below.
Fields enclosed by angle brackets are instantiated from the structured
input specification described in Section~\ref{sec:environment_io_schema}.

\begin{tcolorbox}[
    enhanced,
    breakable,
    colback=gray!3,
    colframe=black!65,
    title=\textbf{System Prompt for Environment Generation},
    fonttitle=\bfseries,
    boxrule=0.7pt,
    arc=2pt,
    left=7pt,
    right=7pt,
    top=5pt,
    bottom=5pt
]
\footnotesize
\ttfamily
You are an environment generator for reinforcement learning training.
Synthesize a standalone Python package under
claw\_envs/<environment\_name>/ that satisfies the following contract.

\textbf{Required artifacts.}

\textbf{1. Environment class.}
Generate a Python class that owns the session lifecycle, orchestrates
domain modules, records actions with deterministic timestamps, and exposes
evaluation entry points. It must implement:

\begin{quote}
create\_session(scenario\_id) -> session\_id\\
reset\_session(session\_id) -> None\\
get\_session\_state(session\_id) -> dict\\
evaluate\_session(session\_id) -> dict
\end{quote}

The returned session state must be compatible with the verifier.
The evaluation method must return per-dimension scores and an
\texttt{overall\_score}.

\textbf{2. Tool functions.}
Generate one Python module for each domain, such as
\texttt{mail.py}, \texttt{social.py}, or \texttt{tickets.py}.
Each module exposes functions that take a session dictionary as input,
return structured dictionaries, and mutate the session state in place.
Domain functions must not parse command-line arguments, acquire file
locks, or access the file system directly.

\textbf{3. State initialization.}
Provide either a module-level \texttt{DEFAULT\_STATE} dictionary or an
\texttt{initial\_session()} factory. Every newly created session must be
initialized from this template. The template must include:

\begin{quote}
session\_id; scenario\_id; created\_at;\\
meta.base\_time; meta.action\_index = 0;\\
workspace\_account; one state container per domain; actions = [].
\end{quote}

\textbf{4. Documentation.}
Generate both \texttt{README.md} and \texttt{SKILL.md}.
The README must describe the task domain, session model, trainer bootstrap
procedure, agent-visible commands, and stress-testing procedure.
The SKILL file must provide a concise English workflow guide, list the
preferred tools and commands, and state critical behavioral constraints.
It must contain fewer than 50 lines.

\textbf{5. Persistence logic.}
Generate a \texttt{store.py} module implementing:

\begin{quote}
load\_session(state\_root, session\_id) -> dict\\
save\_session(state\_root, session\_id, session\_dict) -> None\\
create\_session(state\_root, session\_id, initial\_state) -> None\\
acquire\_lock(state\_root, session\_id) -> Lock\\
release\_lock(lock) -> None
\end{quote}

Session state must be stored at
\texttt{<state\_root>/<session\_id>/session.json}, with the corresponding
lock at \texttt{<state\_root>/<session\_id>/.lock}.
All writes must use an atomic temporary-file-and-replace procedure, and
all reads must remain isolated across sessions.

\textbf{6. Exception handling.}
Every public-facing environment method, CLI command, tool function, and
storage operation must return a structured error dictionary rather than
propagating an exception. Infrastructure failures must be mapped to safe
task-side messages. Returned errors must not expose raw tracebacks,
internal file-system paths, connection strings, or implementation details.
Domain-level validation errors must contain sufficient information for
the agent to correct its action.

\textbf{7. Test cases.}
Generate \texttt{concurrency\_test.py} to validate session isolation,
concurrent-write serialization, successful evaluation of a correct
workflow, penalties for stale or forbidden actions, and correct separation
between hidden trainer commands and agent-visible commands.

\textbf{8. Verifier-compatible state interface.}
The environment class must expose:

\begin{quote}
get\_env\_state() -> dict\\
\_load\_scenario(scenario: dict, long\_context: bool = False) -> None
\end{quote}

The former returns the complete internal state as a plain dictionary.
The latter deep-copies \texttt{DEFAULT\_STATE} and applies scenario
overrides so that the harness can deterministically initialize,
snapshot, and restore the environment.

\textbf{Design constraints.}

One rollout corresponds to one session identifier, and the trainer owns
the complete session lifecycle. The agent must never receive or manually
provide a session identifier. Trainer bindings are injected through the
environment variables
\texttt{<ENV\_NAME>\_SESSION\_ID},
\texttt{<ENV\_NAME>\_STATE\_ROOT}, and
\texttt{<ENV\_NAME>\_SCENARIO\_ID}.

Immutable world data must be stored under \texttt{data/} and must not be
modified during a rollout. Mutable state must be represented as
file-backed JSON with atomic writes and file locks. Timestamps must be
derived from the scenario base time and action index rather than from the
wall clock.

All agent-visible commands must return structured JSON. Hidden trainer
commands, including \texttt{prepare-rollout} and
\texttt{reset-rollout}, must not appear in agent-facing help text.
\end{tcolorbox}

\subsection{Environment Input and Output Schemas}
\label{sec:environment_io_schema}

The environment generator receives a structured specification rather than
an unconstrained natural-language request. The input contains the target
domains, immutable world data, scenario definitions, evaluation rules,
and optional execution constraints.

\paragraph{Input schema.}

\begin{lstlisting}
{
  "environment_name": "string",
  "task_description": "string",
  "domains": ["string"],
  "world_data": {
    "accounts": {
      "<account_id>": {
        "name": "string",
        "role": "string",
        "persona": "string"
      }
    },
    "contacts": {
      "<contact_id>": {
        "name": "string",
        "org": "string",
        "role": "string"
      }
    },
    "scenarios": [
      {
        "scenario_id": "string",
        "task_prompt": "string",
        "base_time": "ISO-8601 string",
        "gold_facts": ["string"],
        "stale_facts": ["string"],
        "required_attachments": ["string"],
        "required_replies": ["string"],
        "allowed_claims": ["string"],
        "forbidden_claims": ["string"],
        "scoring_rules": {
          "<dimension_name>": {
            "weight": "float",
            "rule": "string"
          }
        }
      }
    ],
    "attachments": {
      "<attachment_id>": {
        "filename": "string",
        "content": "string"
      }
    },
    "domain_artifacts": {
      "<domain>": [
        {
          "id": "string",
          "fields": {}
        }
      ]
    }
  },
  "constraints": {
    "max_turns": "optional integer",
    "required_reading_order": ["string"],
    "forbidden_action_sequences": [
      ["string"]
    ]
  }
}
\end{lstlisting}

The \texttt{gold\_facts} field defines information that must be used in
a successful trajectory, whereas \texttt{stale\_facts} contains outdated
or distracting information that must be rejected. Similarly,
\texttt{allowed\_claims} and \texttt{forbidden\_claims} provide
deterministic constraints for the evaluator. The
\texttt{scoring\_rules} field defines the evaluation dimensions and their
weights.

The generator returns the complete package as a structured collection of
files, together with explicit state, command, evaluator, and test
descriptions.

\paragraph{Output schema.}

\begin{lstlisting}
{
  "environment_name": "string",
  "documentation": "README.md content",
  "skill_md": "SKILL.md content",
  "files": [
    {
      "path": "relative package path",
      "content": "complete file content",
      "role": "environment_class | domain_module | cli |
               evaluator | repository | store |
               concurrency_test | doc | data_scenario |
               data_accounts | data_contacts |
               data_attachment | data_artifact"
    }
  ],
  "session_schema": {
    "session_id": "string",
    "scenario_id": "string",
    "created_at": "ISO-8601 string",
    "meta": {
      "base_time": "string",
      "action_index": 0
    },
    "workspace_account": {},
    "<domain>": {},
    "actions": [
      {
        "action_index": 0,
        "timestamp": "string",
        "action_type": "string",
        "details": {}
      }
    ]
  },
  "cli_commands": {
    "hidden": [
      {
        "name": "string",
        "description": "string",
        "bindings_output": ["string"]
      }
    ],
    "agent_visible": [
      {
        "name": "string",
        "description": "string",
        "output_format": {
          "status": "string",
          "data": {}
        }
      }
    ]
  },
  "evaluator": {
    "dimensions": [
      {
        "name": "string",
        "weight": "float",
        "description": "string"
      }
    ],
    "aggregation": "weighted_sum",
    "deterministic": true
  },
  "concurrency_test": {
    "scenarios": ["string"],
    "assertions": ["string"]
  }
}
\end{lstlisting}

The explicit \texttt{role} attached to every generated file enables
automatic completeness checking. An environment is rejected if any
required artifact, interface, or test category is absent.

\subsection{Unified Tool Schema}
\label{sec:tool_schema}

Each generated tool is represented using a unified declarative schema.
Besides conventional names, descriptions, and parameter definitions, the
schema explicitly records whether a tool mutates state and which state
fields may be affected. These annotations are subsequently used to
distinguish read and write tools and to analyze tool dependencies.

\begin{lstlisting}
{
  "name": "string",
  "description": "string",
  "parameters": {
    "type": "object",
    "properties": {
      "<param_name>": {
        "type": "string | integer | number |
                 boolean | array | object",
        "description": "string",
        "default": "optional value"
      }
    },
    "required": ["string"]
  },
  "returns": {
    "type": "object",
    "description": "string",
    "success_keys": ["string"],
    "error_format": {
      "error": "human-readable, agent-safe message"
    }
  },
  "side_effect": "read | write",
  "modifies_state": "boolean",
  "state_fields_affected": ["string"]
}
\end{lstlisting}

A parameter without a \texttt{default} field is treated as required when
it is listed in \texttt{parameters.required}. The
\texttt{side\_effect}, \texttt{modifies\_state}, and
\texttt{state\_fields\_affected} fields provide an explicit operational
description beyond the natural-language tool documentation.

\subsection{Execution and Safety Constraints}
\label{sec:execution_constraints}

The generated package must satisfy the following executable constraints.

\begin{table*}[t]
\centering
\small
\setlength{\tabcolsep}{5pt}
\begin{tabular}{p{0.5cm}p{10.6cm}p{3.2cm}}
\toprule
\textbf{\#} & \textbf{Constraint} & \textbf{Applicable Components} \\
\midrule
1 &
All public functions return dictionaries and map failures to structured
error responses rather than propagating exceptions. &
Environment, tools, CLI, storage, evaluator \\

2 &
No mutable module-level global state is permitted; all mutable data must
reside in the session or environment state. &
All modules \\

3 &
The \texttt{data/} directory is immutable during rollout and can be read
only through the repository layer. &
Repository and domain modules \\

4 &
Session updates use an atomic temporary write followed by
\texttt{os.replace}. &
Storage layer \\

5 &
File locking uses platform-supported locking with a bounded timeout and
an agent-safe error response. &
Storage layer \\

6 &
Timestamps are computed from \texttt{base\_time} and
\texttt{action\_index}; wall-clock timestamps are forbidden. &
Environment and domain modules \\

7 &
CLI outputs follow a structured success or error schema. &
CLI \\

8 &
Agent-visible commands obtain session bindings only from trainer-injected
environment variables and do not expose a \texttt{--session-id} argument. &
CLI \\

9 &
Hidden trainer commands are excluded from agent-facing help messages. &
CLI \\

10 &
Domain modules operate only on the provided session dictionary and do not
perform CLI parsing, file access, or lock acquisition. &
Domain modules \\

11 &
The evaluator is a deterministic, side-effect-free function with no
randomness or external I/O. &
Evaluator \\

12 &
The environment exposes complete state retrieval and deterministic
scenario-loading interfaces for harness snapshot and restoration. &
Environment class \\
\bottomrule
\end{tabular}
\caption{Code-level constraints enforced during environment generation.}
\label{tab:environment_code_constraints}
\end{table*}

Errors are divided into infrastructure failures and domain-level
validation failures. Infrastructure errors are converted into generic
messages that do not expose internal paths or implementation details.
Domain errors identify the invalid entity, missing field, current state,
or valid alternative whenever this information is necessary for agent
self-correction.

Representative error behavior is summarized in
Table~\ref{tab:error_contract}.

\begin{table*}[t]
\centering
\small
\setlength{\tabcolsep}{5pt}
\begin{tabular}{p{3.0cm}p{4.4cm}p{7.0cm}}
\toprule
\textbf{Category} & \textbf{Example} & \textbf{Required Behavior} \\
\midrule
Missing binding &
Session environment variable is absent &
Return an error indicating that the trainer must initialize the rollout. \\

Session not found &
The requested session does not exist &
Return an agent-safe message stating that the session may have expired or
been reset. \\

Lock timeout &
Another worker holds the session lock &
Request a retry without exposing the lock path. \\

Permission failure &
The current account cannot modify an entity &
Identify the affected entity and state that write permission is absent. \\

Invalid transition &
An entity is already in the requested terminal state &
Report the current state and explain why the transition is invalid. \\

Missing field &
A required command argument is absent &
Name the missing field and describe the corresponding requirement. \\

Invalid reference &
A thread, document, or entity identifier does not exist &
Identify the invalid reference and request correction. \\

Stale information &
The agent relies on an outdated artifact &
Allow the tool to emit a warning and let the deterministic evaluator apply
the corresponding penalty. \\

Infrastructure failure &
The state root cannot be accessed &
Return a generic system error while retaining full diagnostic information
only in server-side logs. \\
\bottomrule
\end{tabular}
\caption{Structured error categories used in generated environments.}
\label{tab:error_contract}
\end{table*}

\subsection{Task Generation Prompt}
\label{sec:task_generation_prompt}

For Claw-specific file-workspace tasks, the task generator receives a
sampled operational skeleton consisting of a tool chain, environment
domain, gold actions, and data schema. It then generates four mutually
consistent artifacts in one pass:

\begin{enumerate}[leftmargin=*]
    \item a task configuration file;
    \item a natural user request;
    \item an initial-workspace construction script; and
    \item a deterministic verification script.
\end{enumerate}

The task prompt, workspace state, and verifier must all be derived from a
single internally established ground truth. This one-shot constraint
prevents mismatches in artifact paths, target records, numeric results,
and evaluation conditions.

The core system prompt is given below.

\begin{tcolorbox}[
    enhanced,
    breakable,
    colback=gray!3,
    colframe=black!65,
    title=\textbf{System Prompt for File-Workspace Task Generation},
    fonttitle=\bfseries,
    boxrule=0.7pt,
    arc=2pt,
    left=7pt,
    right=7pt,
    top=5pt,
    bottom=5pt
]
\footnotesize
\ttfamily
You are an AI-agent evaluation architect, scenario writer, and full-stack
engineer. Given a business skeleton consisting of a tool chain and an
environment domain, generate a highly discriminative and purely
code-verifiable file-based workplace task for task
\texttt{<task\_id>}.

Before producing any output, internally establish one ground truth.
Generate the following four files in one pass:

\begin{quote}
tasks/<task\_id>.yaml\\
tasks/prompts/<task\_id>.md\\
tasks/<task\_id>/env\_builder.py\\
scripts/<task\_id>/verify\_workplace.py
\end{quote}

All four files must describe the same objective and unique answer.
The business objective in the user prompt must be derivable from the
workspace generated by \texttt{env\_builder.py}. Every artifact, field,
path, and value checked by \texttt{verify\_workplace.py} must correspond
to an explicit requirement in the user prompt. The verifier must not
introduce a second answer or check an artifact that the prompt does not
request.

Output the four files in the required order. Wrap each file in a Markdown
code block whose first line is its complete relative path. Do not emit
explanations, headings, separators, or scores outside the four code
blocks.

\textbf{Task configuration.}
The YAML file must follow the OpenClaw task schema and include the prompt
path, the asset identifier, and basic runtime configuration.

\textbf{User prompt.}
Write the request as a realistic email, ticket, or verbal handoff from a
specific workplace role. The prompt should communicate the background,
problem, and business objective in the character's voice.

Do not expose hidden answers, expected values, variable names, verifier
logic, scoring rules, or a numbered solution procedure. When referring to
files, use workspace-relative paths only. The request should describe the
desired business outcome while requiring the agent to infer the necessary
operations.

\textbf{Workspace construction.}
Use only the Python standard library. Construct a professional and
non-trivial initial workspace containing distractors, dirty data,
near-duplicate records, outdated versions, decoy entries, and
domain-specific non-standard file formats.

The workspace must nevertheless imply exactly one objectively correct
answer. Distractors may increase reasoning difficulty but must not create
ambiguity. The current working directory is already the task asset
directory, so all paths must be relative. File paths created by the
builder must match those referenced by the prompt and verifier exactly.

\textbf{Deterministic verification.}
The verification script must use only the Python standard library and
must not invoke an LLM, issue network requests, or read mock API
credentials. It should evaluate objective properties such as:

\begin{quote}
exact numeric computation;\\
record extraction, filtering, deduplication, and merging;\\
classification and threshold decisions;\\
set equality and ordering;\\
file or directory structure;\\
schema and format validity.
\end{quote}

Open-ended writing quality must not be used as a primary evaluation
target. When textual output is required, score only structurally
extractable objective elements.

Structured files must be parsed using native JSON, CSV, path, numeric, or
regular-expression operations. Do not use fuzzy substring matching when
exact parsing is available. Unexpected fields or fabricated entities
should receive substantial penalties.

Use a fine-grained weighted score rather than a binary outcome.
The script must write the final result to
\texttt{workplace\_score.json} using the following structure:

\begin{quote}
\{"total\_score": 85,\\
\ \ "details": [\\
\ \ \ \{"item": "...", "score": 10, "max\_score": 10,\\
\ \ \ \ "passed": true, "reason": "..."\}\\
\ \ ]\}
\end{quote}

The verifier receives the workspace path as its first command-line
argument, defaulting to the current directory.

The supplied tool chain, gold actions, environment domain, and data schema
serve only as the underlying business skeleton. Do not mention internal
tool names, CLI commands, or implementation-specific execution commands
in the user-facing task.
\end{tcolorbox}

\subsection{Cross-Artifact Consistency}
\label{sec:cross_artifact_consistency}

The task generator is constrained to produce all task components in a
single response because independent generation can introduce
inconsistencies between the natural-language request, the initial
workspace, and the verifier. EnvCraft therefore enforces the following
cross-artifact invariants:

\begin{enumerate}[leftmargin=*]
    \item \textbf{Path consistency.}
    Every required output path in the user request must exactly match the
    path inspected by the verifier.

    \item \textbf{Data consistency.}
    Every expected value checked by the verifier must be derivable from
    the files generated by the environment builder.

    \item \textbf{Objective consistency.}
    The verifier may assess only outputs and conditions explicitly
    implied by the user request.

    \item \textbf{Answer uniqueness.}
    The initial workspace must admit one objectively correct result,
    despite containing distractors and noisy records.

    \item \textbf{Information separation.}
    Hidden answers, scoring criteria, and verification logic must remain
    absent from the user-facing prompt.

    \item \textbf{Code-only evaluation.}
    The final reward is computed by deterministic programmatic checks
    rather than by an LLM-based judge.
\end{enumerate}

Together, the explicit environment contract, unified tool schema, and
cross-artifact task-generation protocol make generated samples directly
executable and deterministically verifiable. They also enable automated
auditing of missing package components, invalid tool interfaces,
non-isolated state mutations, inconsistent artifact paths, and
non-deterministic reward logic.


\section{Experimental Results}
\subsection{Training Hyperparameters}
\label{sec:training_hyperparameters}

We train all models using Group Relative Policy Optimization (GRPO)
implemented with the VERL framework~\citep{sheng2024verl}.
Unless otherwise specified, all experiments use the same optimization,
rollout, and evaluation configurations. The detailed hyperparameters
are summarized in Table~\ref{tab:training_hyperparameters}.
\begin{table}[t]
    \centering
    \small
    \setlength{\tabcolsep}{6pt}
    \begin{tabular}{@{}ll@{}}
        \toprule
        \textbf{Hyperparameter} & \textbf{Value} \\
        \midrule

        \multicolumn{2}{@{}l}{\textit{\textbf{Optimization}}} \\
        Optimizer & AdamW \\
        Learning rate & $1\times10^{-6}$ \\
        Learning-rate scheduler & Constant \\
        Warmup ratio & $0.0$ \\
        Weight decay & $0.01$ \\
        Gradient clipping & $1.0$ \\
        Training epochs & 1 \\
        Global batch size & 64 \\

        \midrule
        \multicolumn{2}{@{}l}{\textit{\textbf{GRPO}}} \\
        Group size $G$ & 8 \\
        Clipping coefficient $\epsilon_{\mathrm{clip}}$ & $0.2$ \\
        KL coefficient $\beta$ & $0.02$ \\
        Advantage normalization & Group-wise normalization \\
        Entropy coefficient & $0.0$ \\

        \midrule
        \multicolumn{2}{@{}l}{\textit{\textbf{Rollout}}} \\
        Rollout batch size & 64 \\
        Rollouts per prompt & 8 \\
        Temperature & $1.0$ \\
        Top-$p$ & $1.0$ \\
        Maximum prompt length & 4,096 \\
        Maximum generation length & 32,768 \\
        Maximum action turns & 32 \\

        \midrule
        \multicolumn{2}{@{}l}{\textit{\textbf{Infrastructure}}} \\
        Number of GPUs & 64 NPUs ($8$ nodes $\times$ $8$ NPUs) \\
        GPU type & Huawei Ascend 910B NPU \\
        Tensor parallel size & 4 \\
        Data parallel size & 8 rollout replicas \\
        Gradient checkpointing & Enabled \\

        \bottomrule
    \end{tabular}
    \caption{
    Common hyperparameters used for EnvCraft reinforcement learning.
    Model-specific differences are described in the main text.
    }
    \label{tab:training_hyperparameters}
\end{table}
\subsection{Training Data and Curriculum}
\label{sec:training_curriculum}

We consider two training strategies. In direct Claw-specific training,
the model is optimized solely on the EnvCraft-Claw dataset. In the
two-stage Tool-to-Claw curriculum, the model is first trained on
EnvCraft-Tool and subsequently optimized on EnvCraft-Claw.

For the first stage, we train the model for \texttt{115} steps using
\texttt{8k} EnvCraft-Tool samples. The resulting checkpoint is then
used to initialize the second stage, which is trained for
\texttt{30} steps using \texttt{2k} EnvCraft-Claw samples.
The learning rate is \texttt{$1\times10^{-6}$} in the first stage and
\texttt{$1\times10^{-6}$} in the second stage. All other
hyperparameters are shared across the two stages.

\subsection{Reward Assignment and Rollout Handling}
\label{sec:reward_assignment}
Each rollout receives an outcome reward from the programmatic verifier
associated with its training sample. For EnvCraft-Tool, we adopt a
binary reward: a successfully completed trajectory receives a reward of
$1.0$, whereas an unsuccessful trajectory receives a reward of $0.0$.
For EnvCraft-Claw, the verifier assigns a normalized reward in the
range $[0,1]$ according to the degree of task completion, where $1.0$
indicates full completion and $0.0$ indicates complete failure.
Intermediate values represent partial task completion. Unless otherwise
specified, no intermediate process-level reward is assigned.

A rollout is considered unsuccessful if the agent produces an invalid
tool call, exceeds the maximum number of action turns, reaches the
execution timeout, triggers an unrecoverable environment exception, or
fails to make verifiable progress toward the target state. Recoverable
tool-execution errors remain part of the interaction trajectory and do
not immediately terminate the rollout, allowing the policy to learn
error-recovery behavior. The final reward is assigned after the
resulting environment state is evaluated by the corresponding
deterministic verifier.
Verifier execution errors are treated as failed verification and the
corresponding rollout is assigned a reward of $0.0$.

\subsection{Evaluation Configuration}
\label{sec:evaluation_configuration}

For evaluation, all models are decoded using a temperature of
\texttt{1.0}, top-$p$ of \texttt{0.6}, and a maximum generation
length of \texttt{32768} tokens. Each task permits at most
\texttt{32} agent turns and has an execution timeout of
\texttt{300 seconds}. We use \texttt{8} independent runs per model and
report the \texttt{mean} score. The reported error terms denote
\texttt{standard deviation} across
\texttt{evaluation runs}.

For PinchBench and Claw-Eval, task success is determined using the
official evaluation protocol. For BFCL-v3 and $\tau^2$-bench, we follow
the benchmark-provided evaluation scripts and default task splits.
All models use the same decoding and execution settings unless the
official model interface imposes different constraints.

\subsection{Training Dynamics and Data Composition}
\begin{figure*}[t]
    \centering

    \begin{subfigure}[b]{0.32\textwidth}
        \centering
        \includegraphics[width=\linewidth]
        {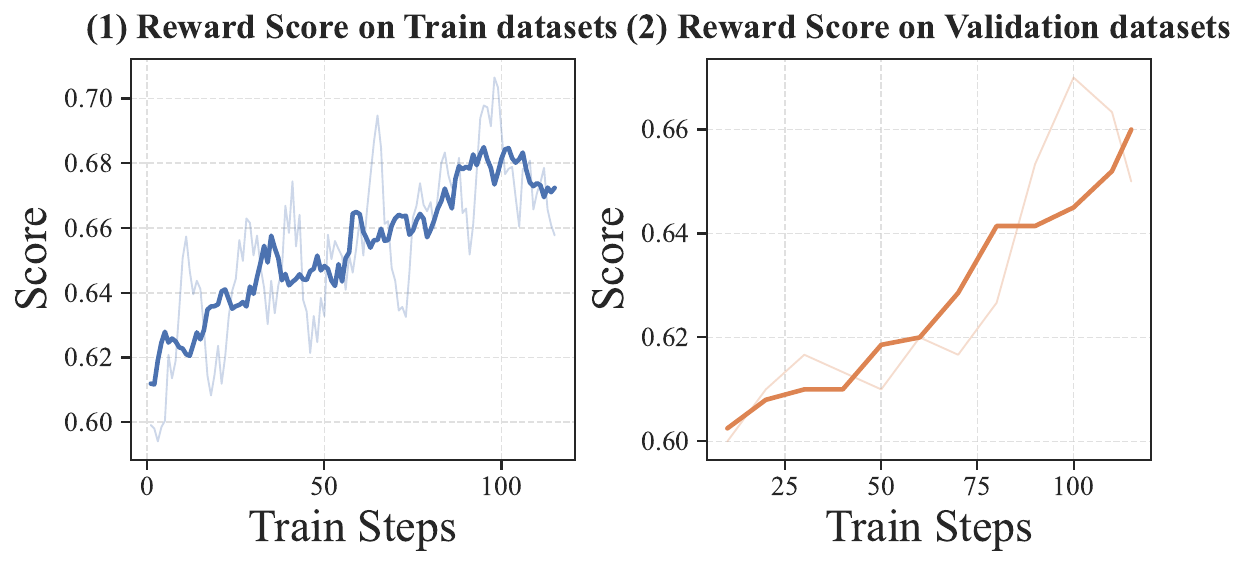}
        \caption{Qwen3-8B}
        \label{fig:tool_curve_qwen3_8b}
    \end{subfigure}
    \hfill
    \begin{subfigure}[b]{0.32\textwidth}
        \centering
        \includegraphics[width=\linewidth]
        {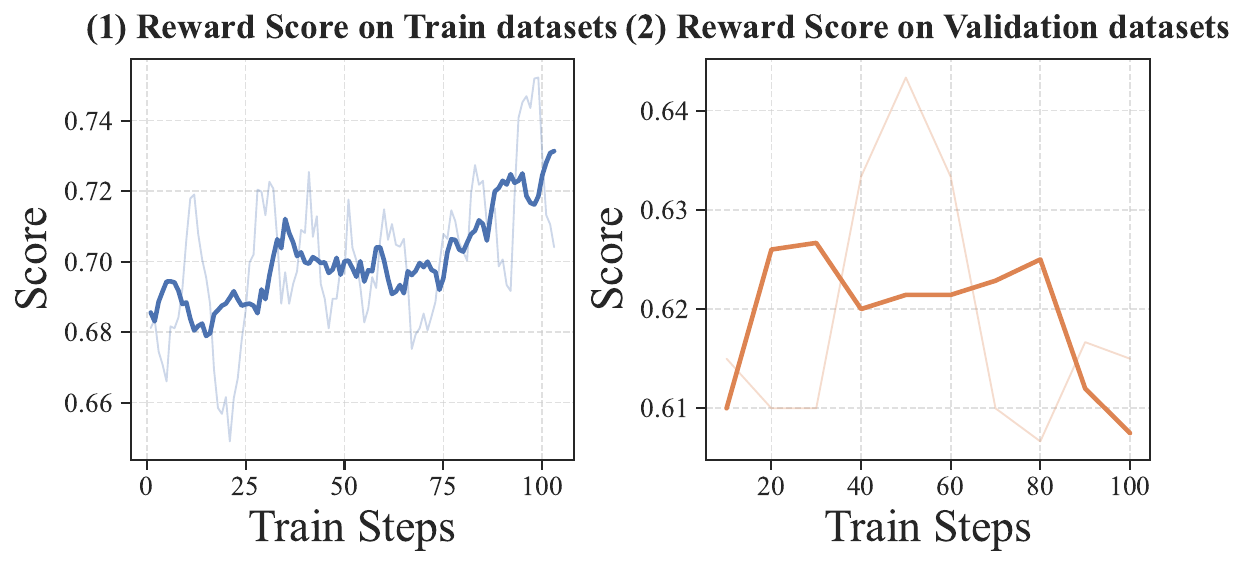}
        \caption{Qwen3-32B}
        \label{fig:tool_curve_qwen3_32b}
    \end{subfigure}
    \hfill
    \begin{subfigure}[b]{0.32\textwidth}
        \centering
        \includegraphics[width=\linewidth]
        {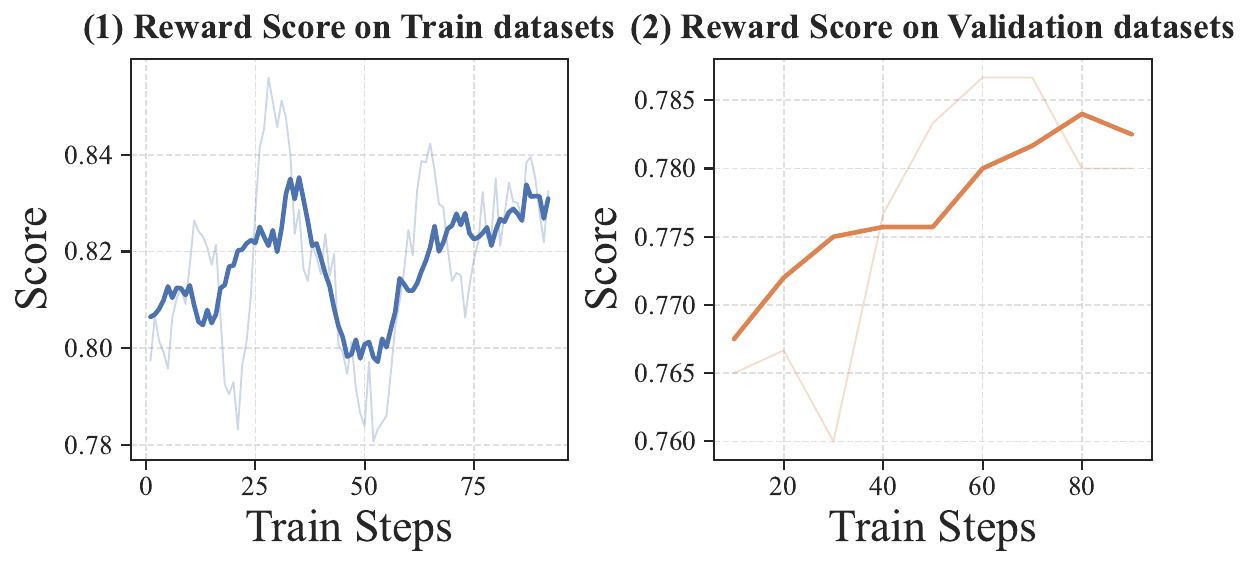}
        \caption{Qwen3.5-9B}
        \label{fig:tool_curve_qwen35_9b}
    \end{subfigure}

    \caption{
Training dynamics of Qwen3-8B, Qwen3-32B, and Qwen3.5-9B on
EnvCraft-Tool. The left panels show training performance, while the
right panels report performance on 50 held-out tasks excluded from
training.
}
    \label{fig:tool_training_dynamics}
\end{figure*}

\paragraph{Training Dynamics on EnvCraft-Tool.}
We examine the training dynamics of Qwen3-8B, Qwen3-32B, and
Qwen3.5-9B on EnvCraft-Tool. As shown in
Figure~\ref{fig:tool_training_dynamics}, the left panels report training
performance, while the right panels evaluate intermediate checkpoints
on a held-out set of 50 EnvCraft-Tool tasks reserved before training
and excluded from optimization. All three backbones exhibit overall
improvements on both splits, despite checkpoint-level fluctuations,
indicating that EnvCraft-Tool provides consistent learning signals
rather than merely encouraging memorization of training instances.

\begin{figure*}[t]
    \centering

    \begin{subfigure}[b]{0.32\textwidth}
        \centering
        \includegraphics[width=\linewidth]
        {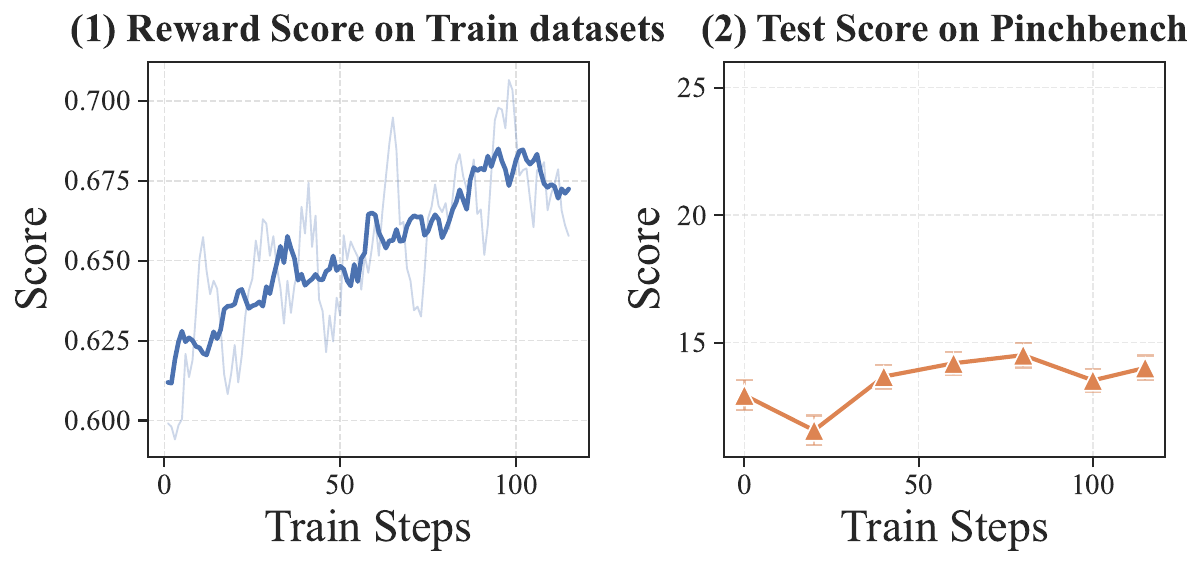}
        \caption{General Tool-use training}
    \end{subfigure}
    \hfill
    \begin{subfigure}[b]{0.32\textwidth}
        \centering
        \includegraphics[width=\linewidth]{./qwen3_8b_claw_3.pdf}
        \caption{Claw-specific training}
    \end{subfigure}
    \hfill
    \begin{subfigure}[b]{0.32\textwidth}
        \centering
        \includegraphics[width=\linewidth]
        {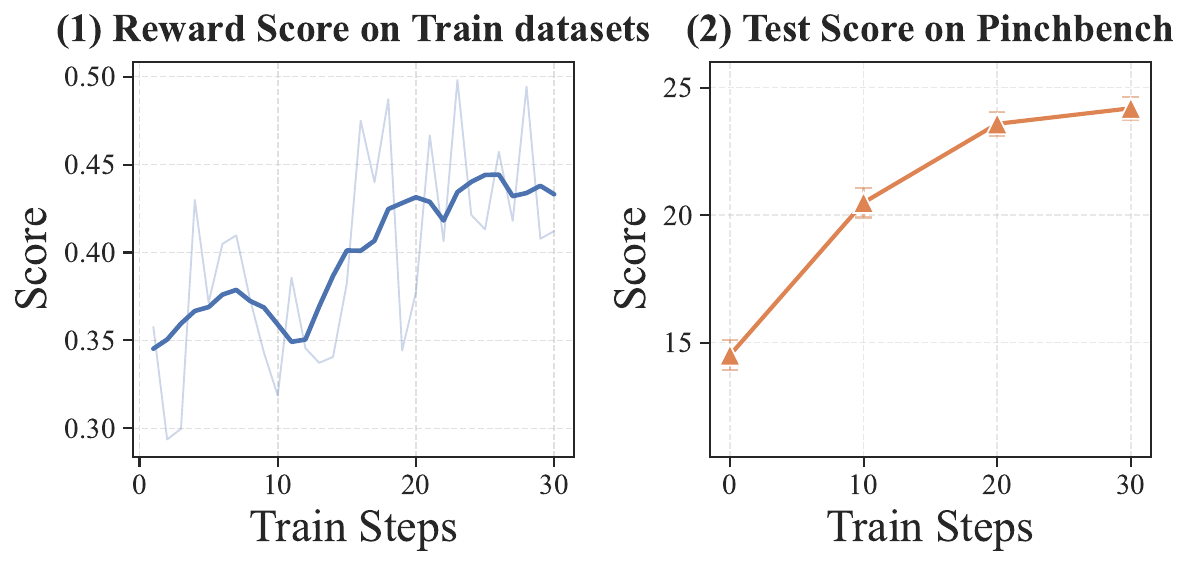}
        \caption{Tool-to-Claw training}
    \end{subfigure}
    \caption{
    Effect of training-data composition on Qwen3-8B. We compare general
    tool-use training, Claw-specific training, and sequential
    general-to-Claw training using checkpoint-level performance on
    PinchBench.
    }
    \label{fig:data_composition}
\end{figure*}

\paragraph{Effect of Training Data Composition.}
We investigate how different data compositions affect claw-task performance on Qwen3-8B (Fig.~\ref{fig:data_composition}). 
General tool-use training yields modest gains: PinchBench rises +1.57\% and Claw-Eval rises +1.74\%. 
In contrast, Claw-specific training delivers substantially larger improvements, reaching +11.91\% and +11.36\%, respectively. 
Sequential Tool-to-Claw training achieves comparable results (24.19 and 55.63). 
These results indicate that domain-aligned Claw trajectories, which provide stateful workspaces, long-horizon execution, and cross-tool dependencies, contribute the majority of system-level gains. 
General tool-use data offers complementary supervision for basic function selection and parameter grounding, yet its standalone effect is limited. 
Overall, the findings suggest that task-aligned executable trajectories are more critical than simply accumulating heterogeneous training data.

\begin{figure*}[t]
    \centering

    \begin{subfigure}[b]{0.48\textwidth}
        \centering
        \includegraphics[width=\linewidth]{./bfcl_base_scores.pdf}
        \caption{General tool-use training}
    \end{subfigure}
    \hfill
    \begin{subfigure}[b]{0.48\textwidth}
        \centering
        \includegraphics[width=\linewidth]{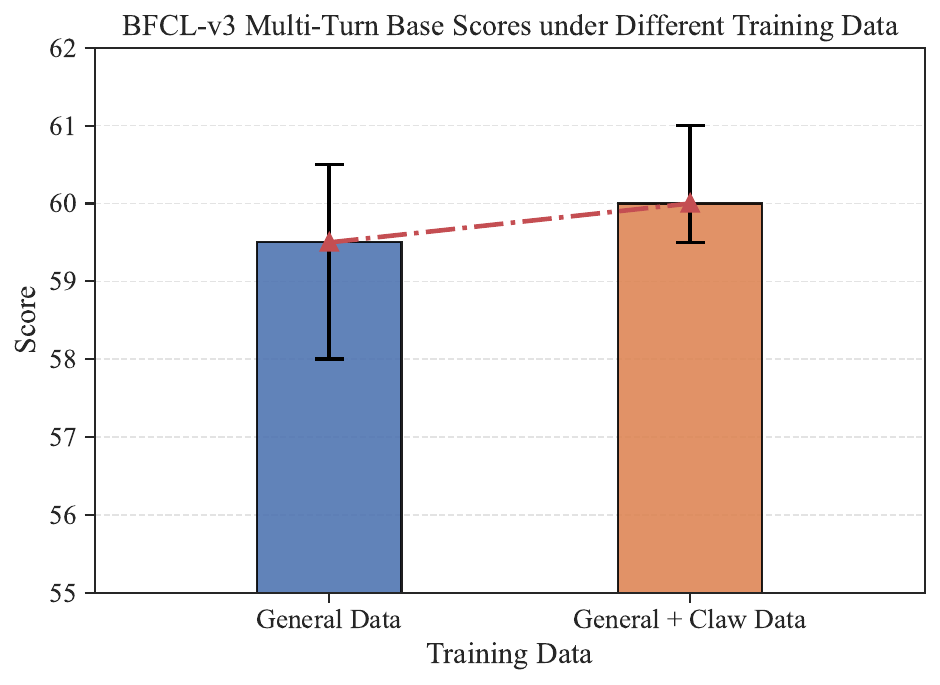}
        \caption{Continual Claw-specific training}
    \end{subfigure}

    \caption{
    Retention of general tool-use capability during Claw-specific
    reinforcement learning. Scores are measured on the BFCL-v3
    Multi-Turn Base subset at different training checkpoints.
    }
    \label{fig:capability_retention}
\end{figure*}

\paragraph{Retention of General Tool-Use Capability.}
A potential concern is that specializing a model on Claw-specific environments may degrade its general tool-use capability. To examine this issue, we evaluate intermediate and final checkpoints on the BFCL-v3 Multi-Turn Base subset. Figure~\ref{fig:capability_retention} compares general tool-use training with subsequent Claw-specific training. After switching to Claw-specific environments, the model maintains a BFCL score of 60.0, compared with 59.5 before the transition, while its PinchBench performance improves from 14.51 to 24.19. These results provide evidence that Claw-specific reinforcement learning
improves system-level execution without measurable degradation on the
BFCL-v3 Multi-Turn Base subset.
\subsection{Data Quality Analysis}
\paragraph{LLM-Based Quality-Auditing Rubric.}
We employ an LLM-based inspector to assess the logical validity,
executability, and internal consistency of each synthesized task before
difficulty calibration. For each candidate, the inspector receives four
inputs: the complete environment source code, the initialized environment
state, the user intent sequence, and the corresponding programmatic
validation protocol. The inspector then evaluates the candidate along
three dimensions: task solvability and logic, validation-protocol
integrity, and cross-component data consistency.

Specifically, we use \textbf{DeepSeek-V4-Pro} as the auditing model.
Each candidate is inspected once using deterministic decoding, and the
three dimension scores returned by the inspector are directly used for
quality filtering. In addition to numerical scores, the inspector
produces a concise, dimension-specific justification to facilitate
failure analysis. Table~\ref{tab:llm_auditing_rubric} presents the
complete scoring rubric.

\begin{table*}[t]
    \centering
    \small
    \setlength{\tabcolsep}{4pt}
    \renewcommand{\arraystretch}{1.12}
    \begin{tabular}{
        p{2.35cm}
        p{3.25cm}
        p{3.25cm}
        p{3.25cm}
        p{3.25cm}
    }
        \toprule
        \textbf{Dimension}
        & \textbf{Score 3}
        & \textbf{Score 2}
        & \textbf{Score 1}
        & \textbf{Score 0} \\
        \midrule

        \textbf{Task Solvability \& Logic}
        &
        The task is logically valid, non-trivial, and fully solvable
        under the initial state and environment rules. Completing it
        requires meaningful interaction, with no state conflicts.
        &
        The task is theoretically solvable but is either trivial,
        requires no meaningful interaction, or contains minor parameter
        mismatches that a robust agent may bypass.
        &
        The task is unsolvable because its requested actions or target
        state contradict the initial state, environment constants, or
        transition rules.
        &
        The task substantially hallucinates entities, tools, APIs, or
        states that are absent from the environment.
        \\

        \midrule

        \textbf{Validation-Protocol Integrity}
        &
        The validation protocol is executable and completely checks the
        intended success conditions, without evident false-positive or
        false-negative paths.
        &
        The protocol is executable, but its assertions are incomplete or
        weak; for example, it verifies that an action occurred without
        checking the required final state.
        &
        The protocol contains syntax or runtime errors, references
        undefined variables or APIs, or is expected to fail during
        execution.
        &
        The validation protocol is missing or evaluates conditions
        unrelated to the user task.
        \\

        \midrule

        \textbf{Data Consistency \& Syntactic Cleanliness}
        &
        Entity names, identifiers, value types, and schemas are fully
        consistent across the environment state, user task, and
        validation protocol.
        &
        The components are semantically consistent but contain minor
        type or formatting mismatches that do not fundamentally alter
        the task objective.
        &
        Critical fields are missing, or explicit contradictions exist
        between the user-facing description and the underlying
        environment configuration.
        &
        The task record contains malformed JSON, severely corrupted
        text, or otherwise unusable structured data.
        \\

        \bottomrule
    \end{tabular}
    \caption{
    Complete rubric used by the LLM-based data-quality inspector.
    Each synthesized task is assigned an integer score from 0 to 3
    along the three dimensions.
    }
    \label{tab:llm_auditing_rubric}
\end{table*}

Let
\(
s_{\mathrm{sol}},
s_{\mathrm{val}},
s_{\mathrm{con}}
\in \{0,1,2,3\}
\)
denote the scores for task solvability, validation integrity, and data
consistency, respectively. A candidate is retained only if

\begin{equation}
\small
\begin{aligned}
s_{\mathrm{sol}} &= 3, \\
s_{\mathrm{val}} &= 3, \\
s_{\mathrm{con}} &\geq 2, \\
s_{\mathrm{sol}} + s_{\mathrm{val}} + s_{\mathrm{con}}
&\geq 8.
\end{aligned}
\label{eq:quality_gate}
\end{equation}

This strict gate requires every retained task to be fully solvable and
equipped with a reliable programmatic reward function, while permitting
only minor formatting inconsistencies that do not affect execution.
Candidates failing any of these conditions are removed before the
subsequent rollout-based difficulty calibration stage.

The auditing prompt provides the inspector with the raw environment code,
initial state, user intent sequence, and validation script. It explicitly
defines the score anchors in Table~\ref{tab:llm_auditing_rubric} and
requires the inspector to return only a structured JSON object containing
dimension-specific analyses, numerical scores, and the final qualification
decision. A compact representation of the required output format is shown
below.

\begin{lstlisting}[
    caption={Structured output schema of the LLM-based quality inspector.},
    label={lst:quality_auditing_output}
]
{
  "reasoning_steps": {
    "dimension_1_analysis":
      "Evidence-based analysis of task solvability.",
    "dimension_2_analysis":
      "Evidence-based analysis of validation integrity.",
    "dimension_3_analysis":
      "Evidence-based analysis of data consistency."
  },
  "scores": {
    "task_solvability": 0,
    "validation_integrity": 0,
    "data_consistency": 0
  },
  "is_qualified": false
}
\end{lstlisting}

\begin{figure}[t]
    \centering
    \includegraphics[width=\linewidth]
    {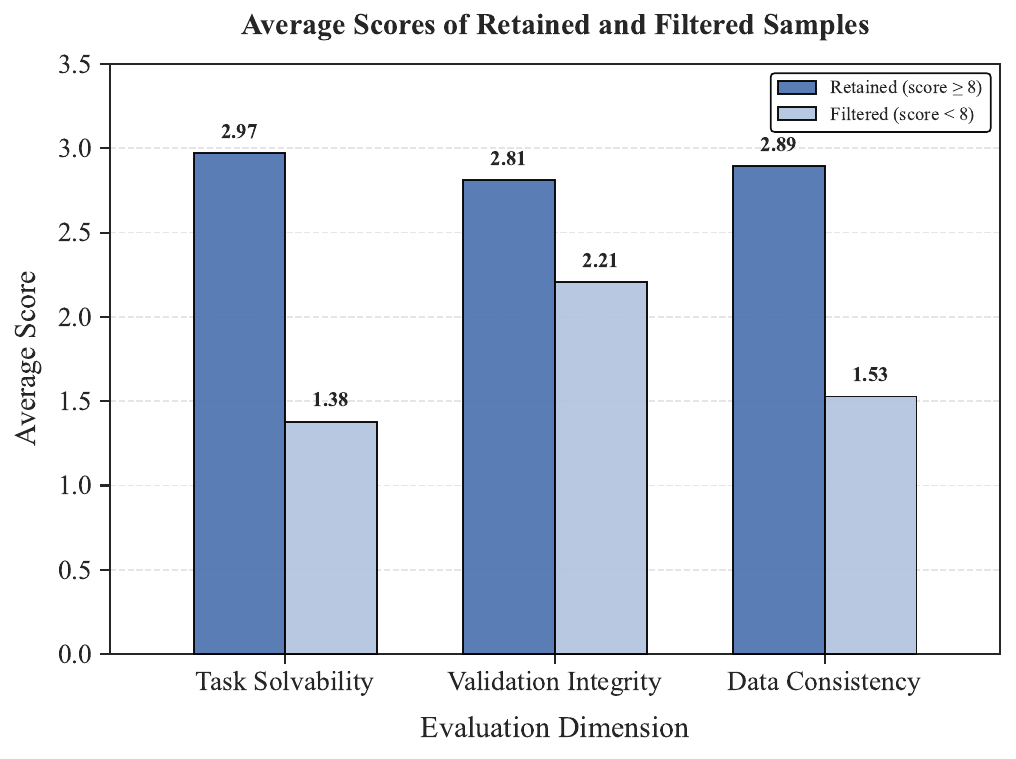}
    \caption{
    Average LLM-rubric scores of retained and filtered task candidates
    across task solvability, validation-protocol integrity, and data
    consistency. Each dimension is scored on a 0--3 scale. A candidate
    is retained only when it receives scores of 3, 3, and at least 2
    on the three dimensions, respectively.
    }
    \label{fig:rubric_analysis}
\end{figure}

\paragraph{Effectiveness of LLM-Based Quality Auditing.}
We further examine whether the LLM-based inspector meaningfully
distinguishes high-quality tasks from invalid candidates.
Figure~\ref{fig:rubric_analysis} compares retained and filtered samples
in terms of their average task-solvability, validation-integrity, and
data-consistency scores.

Retained samples achieve consistently higher scores across all three
dimensions, with the largest separation observed in task solvability.
In particular, the mean rubric score, averaged across the three
dimensions, increases from 1.38 for filtered candidates to 2.97 for
retained samples. Because retained candidates must satisfy the strict
conditions in Equation~\ref{eq:quality_gate}, their scores are
concentrated near the upper end of the rubric. These results indicate
that the auditing procedure does not merely enforce syntactic
correctness, but also filters tasks with contradictory execution logic,
inconsistent environment states, or unreliable reward functions.

\paragraph{Difficulty Calibration.}
After rubric-based filtering, we calibrate task difficulty by evaluating
each candidate task with Qwen3.5-27B over eight independent rollouts.
All rollouts use the same system prompt, user instruction, tool
definitions, and initial environment state, while using different random
seeds to induce independent trajectories. We adopt a sampling temperature
of $0.7$ and a top-$p$ value of $0.95$ for all rollouts.

Let $z_j \in \{0,1\}$ denote the outcome of the $j$-th rollout. A rollout
is considered successful ($z_j=1$) only if it terminates within the
maximum interaction budget and the deterministic verifier confirms that
the resulting environment state satisfies all task constraints. A rollout
is considered unsuccessful ($z_j=0$) if the verifier returns a negative
result, the agent reaches the maximum of 32 action turns without
completing the task, or the rollout exceeds a wall-clock timeout of
300 seconds. The empirical pass rate of a task is computed as
\begin{equation}
p = \frac{1}{8}\sum_{j=1}^{8} z_j.
\end{equation}

We retain only tasks satisfying
\begin{equation}
0 < p < 1,
\end{equation}
corresponding to tasks solved in one to seven of the eight rollouts.
Tasks with $p=1$ are removed because they provide limited learning
signals, whereas tasks with $p=0$ are excluded because they are likely
unsolvable, underspecified, or excessively difficult for the current
training stage.

We distinguish verifier rejection from verifier failure. A verifier that
executes normally but returns a negative result is treated as a failed
rollout. In contrast, verifier runtime exceptions, malformed outputs, or
verifier timeouts are treated as data-quality defects rather than agent
failures. Such samples are removed from difficulty calibration and routed
to the verifier-repair and re-auditing pipeline. We allow at most two
automatic repair attempts and discard samples whose verifiers remain
invalid afterward.

\subsection{Ablation Studies}
\begin{figure}[t]
    \centering
    \includegraphics[width=\linewidth]
    {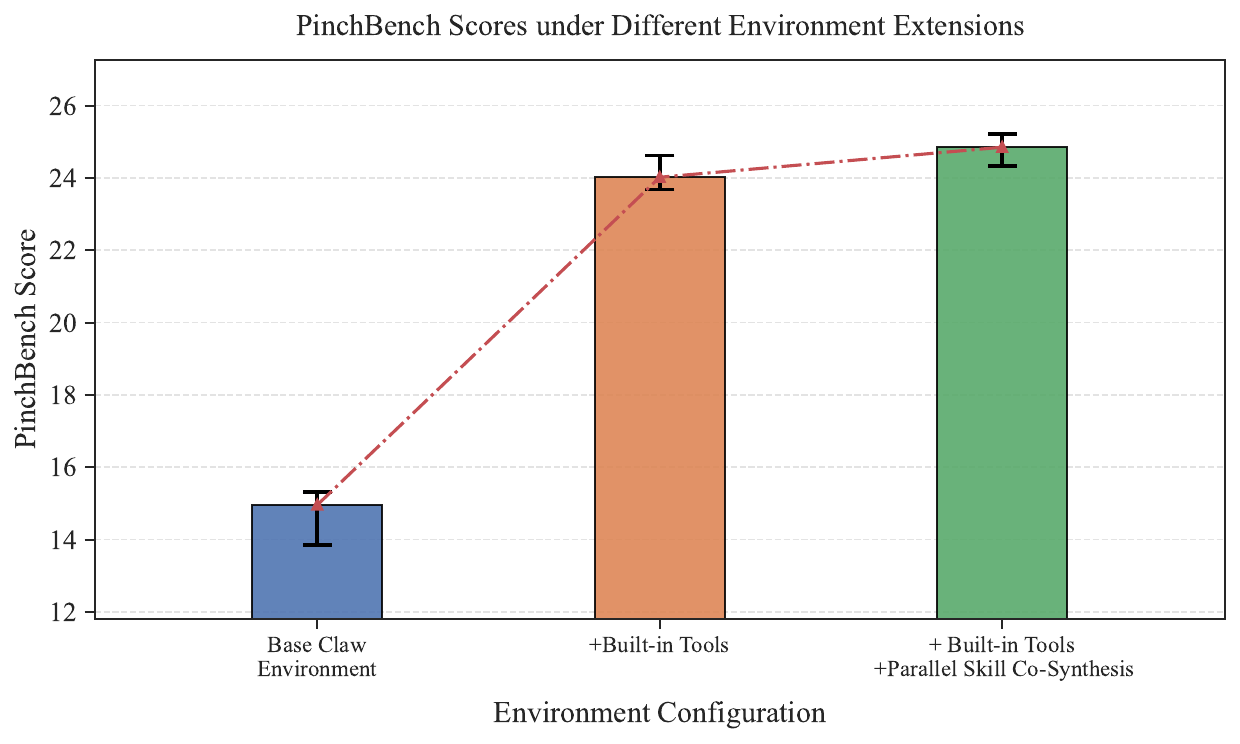}
    \caption{
    Ablation study of Claw-specific environment components on PinchBench.
The three configurations progressively introduce built-in system tools
and parallel skill co-synthesis. All models use the same Qwen3-8B
backbone and training configuration.
    }
    \label{fig:environment_ablation}
\end{figure}

\paragraph{Contribution of Claw-Specific Environment Extensions.}
To examine the contribution of the Claw-specific environment design, we train Qwen3-8B under three progressively enhanced configurations and evaluate the resulting models on PinchBench. As shown in Figure~\ref{fig:environment_ablation}, augmenting the base Claw environment with built-in system-level tools substantially improves the score from 14.96 to 24.02, demonstrating that direct access to executable workspace operations provides the primary learning signal for system-level agent capabilities. Further introducing parallel skill co-synthesis increases the score to 24.33, yielding a smaller but consistent additional gain. These results suggest that built-in tools are the dominant contributor, while synthesized skills provide complementary workflow guidance that helps agents more effectively utilize the available execution interfaces. The best performance is achieved when both components are combined, validating the effectiveness of the complete Claw-specific environment design.

\section{Qualitative Trajectory Example}
\label{sec:qualitative_trajectory}

\paragraph{Cross-File Performance Aggregation.}
We present a representative EnvCraft-Claw trajectory involving
multi-file workspace inspection, structured data integration, and
programmatic state mutation. The agent must identify valid records
across an employee roster, a monthly output ledger, and role-specific
scoring rules, while excluding incomplete or inconsistent entries.
It then computes weighted performance scores, writes the results to a
designated workspace file, and verifies the generated artifact.
This example illustrates how EnvCraft tasks require agents to combine
information retrieval, constraint checking, numerical computation, and
persistent file operations within a single executable workflow.
For readability, we omit repetitive intermediate reasoning and
abbreviate lengthy tool outputs while preserving the original action
sequence and execution results.
\begin{figure*}[p]
\centering

\begin{tcolorbox}[
    enhanced,
    width=\textwidth,
    colback=gray!3,
    colframe=black!65,
    title=\textbf{Representative EnvCraft-Claw Trajectory:
    Performance Score Aggregation},
    fonttitle=\bfseries,
    boxrule=0.7pt,
    arc=2pt,
    left=8pt,
    right=8pt,
    top=6pt,
    bottom=6pt
]

\small
\setlength{\parindent}{0pt}
\setlength{\parskip}{2pt}

\textbf{User Request.}
Linda, an HR specialist, asks the agent to prepare a monthly
performance report. The workspace contains an employee roster under
\texttt{data/employees/}, monthly output records under
\texttt{data/ledgers/}, and role-specific scoring rules under
\texttt{data/rules/}. Some records may be incomplete or redundant.
The agent must retain only employees who:
(1) appear in the roster,
(2) have a monthly output record, and
(3) have a role with a defined scoring rule.
It must compute their weighted scores and save the resulting report to
\texttt{performance/scores\_202501.json}.

\medskip
\hrule
\medskip

\textbf{Step 1: Inspect the workspace}

\textit{Agent rationale.}
First, identify the available data sources and confirm that the target
output directory exists.

\textbf{Tool calls:}
\begin{quote}
\texttt{ls(path="data")}\\
\texttt{ls(path="performance")}
\end{quote}

\textbf{Observation.}
The workspace contains three data directories:
\texttt{employees}, \texttt{ledgers}, and \texttt{rules}.
The \texttt{performance} directory is currently empty.

\medskip
\hrule
\medskip

\textbf{Step 2: Load the relevant records}

\textit{Agent rationale.}
Read the employee roster, monthly output ledger, and scoring-rule files
to determine their schemas and join keys.

\textbf{Tool calls:}
\begin{quote}
\texttt{read("data/employees/employees.json")}\\
\texttt{read("data/ledgers/monthly\_outputs.json")}\\
\texttt{read("data/rules/scoring\_rules.json")}
\end{quote}

\textbf{Observation.}
The files use \texttt{employee\_id} as the employee-level join key and
\texttt{role\_code} as the rule-level join key. Scoring weights are
available for Software Engineers (\texttt{SE}) and Quality Assurance
staff (\texttt{QA}), but not for Human Resources staff (\texttt{HR}).

\medskip
\hrule
\medskip

\textbf{Step 3: Resolve inconsistent records}

\textit{Agent rationale.}
Cross-reference the three sources and retain only records satisfying all
eligibility conditions.

\begin{itemize}[
    leftmargin=1.6em,
    itemsep=1pt,
    topsep=2pt,
    parsep=0pt
]
    \item \texttt{emp001}, \texttt{emp002}, and \texttt{emp003} are
    valid across all three sources.
    
    \item \texttt{emp004} is excluded because the \texttt{HR} role has
    no scoring rule.
    
    \item \texttt{emp005} is excluded because no monthly output record
    is available.
    
    \item \texttt{emp999} is excluded because it does not appear in the
    employee roster.
\end{itemize}

For each eligible employee, the agent computes

\[
\begin{aligned}
s ={}&
w_{\mathrm{delivery}}x_{\mathrm{delivery}}
+
w_{\mathrm{quality}}x_{\mathrm{quality}}
\\
&+
w_{\mathrm{collaboration}}
x_{\mathrm{collaboration}}.
\end{aligned}
\]

\end{tcolorbox}

\caption{
A representative EnvCraft-Claw trajectory involving workspace
inspection, cross-file record matching, rule-based computation,
persistent state mutation, and output verification.
For readability, repetitive intermediate reasoning and lengthy tool
outputs are omitted while preserving the original action sequence and
execution results.
}
\label{fig:representative_tool_trajectory}

\end{figure*}

\begin{figure*}[p]
\ContinuedFloat
\centering

\begin{tcolorbox}[
    enhanced,
    width=\textwidth,
    colback=gray!3,
    colframe=black!65,
    title=\textbf{Representative EnvCraft-Tool Trajectory:
    Performance Score Aggregation (Continued)},
    fonttitle=\bfseries,
    boxrule=0.7pt,
    arc=2pt,
    left=8pt,
    right=8pt,
    top=6pt,
    bottom=6pt
]

\small
\setlength{\parindent}{0pt}
\setlength{\parskip}{2pt}

\textbf{Step 4: Generate the performance report}

\textit{Agent rationale.}
Use a Python script to join the records, calculate the weighted scores,
sort employees by total score, and write a structured JSON artifact.

\textbf{Tool call:}
\begin{quote}
\texttt{execute\_bash(python scoring script)}
\end{quote}

\textbf{Observation.}
Three eligible employees are identified:

\begin{center}
\setlength{\tabcolsep}{10pt}
\renewcommand{\arraystretch}{1.12}

\begin{tabular}{@{}lcc@{}}
\toprule
\textbf{Employee} & \textbf{Role} & \textbf{Weighted Score} \\
\midrule
Alice (\texttt{emp001})   & SE & 92.0 \\
Bob (\texttt{emp002})     & SE & 84.5 \\
Charlie (\texttt{emp003}) & QA & 80.0 \\
\bottomrule
\end{tabular}
\end{center}

The results are written to
\texttt{performance/scores\_202501.json}.

\medskip
\hrule
\medskip

\textbf{Step 5: Verify the workspace mutation}

\textit{Agent rationale.}
Read the generated file to verify that the requested artifact exists and
contains the expected records and summary statistics.

\textbf{Tool call:}
\begin{quote}
\texttt{read("performance/scores\_202501.json")}
\end{quote}

\textbf{Observation.}
The file contains three employee records, with two \texttt{SE} employees
and one \texttt{QA} employee. The scores and ordering match the
role-specific weighted calculations.

\medskip
\hrule
\medskip

\textbf{Final Response.}
The performance report has been generated successfully and saved to
\texttt{performance/scores\_202501.json}. Three employees satisfy all
eligibility conditions: Alice with a score of 92.0, Bob with 84.5, and
Charlie with 80.0. Diana is excluded because no HR scoring rule exists,
Eve because her monthly ledger is missing, and \texttt{emp999} because
the identifier is absent from the employee roster.

\end{tcolorbox}

\caption{
Representative EnvCraft-Claw trajectory continued.
}
\end{figure*}
